\documentclass[10pt,twocolumn,letterpaper]{article}

\usepackage[pagenumbers]{cvpr} %

\usepackage{multirow}
\usepackage{subcaption}

\usepackage[dvipsnames]{xcolor}

\usepackage{times}
\usepackage{epsfig}
\usepackage{graphicx}
\usepackage{amsmath}
\usepackage{amssymb}
\usepackage{picture}
\usepackage{booktabs}
\usepackage{caption}
\usepackage{fontawesome}
\usepackage{multirow}
\usepackage{subcaption}
\usepackage{bm}
\usepackage{bbm}
\usepackage{tikz}
\usepackage{xcolor}
\usepackage{wrapfig}

\definecolor{cvprblue}{rgb}{0.21,0.49,0.74}
\usepackage[pagebackref,breaklinks,colorlinks,citecolor=cvprblue]{hyperref}

\renewcommand*{\ie}{i.e.,\@\xspace}
\renewcommand*{\eg}{e.g.,\@\xspace}

\def\paperID{6} %
\def\confName{CVPR}
\def\confYear{2024}

\begin{document}

\title{Enhancing 2D Representation Learning with a 3D Prior}

\author{Mehmet Ayg\"un$^{1}$\thanks{Work done during MA's internship at Meta Inc.}\qquad Prithviraj Dhar$^{2}$ \qquad Zhicheng Yan $^{2}$ \qquad Oisin Mac Aodha$^{1}$ \qquad Rakesh Ranjan$^{2}$\qquad  \\[4pt]
$^{1}$ University of Edinburgh \quad
$^{2}$ Meta Inc.
}

\maketitle
\begin{abstract}
Learning robust and effective representations of visual data is a fundamental task in computer vision. Traditionally, this is achieved by training models with labeled data which can be expensive to obtain. Self-supervised learning attempts to circumvent the requirement for labeled data by learning representations from raw unlabeled visual data alone. However, unlike humans who obtain rich 3D information from their binocular vision and through motion, the majority of current self-supervised methods are tasked with learning from monocular 2D image collections. This is noteworthy as it has been demonstrated that shape-centric visual processing is more robust compared to texture-biased automated methods. Inspired by this, we propose a new approach for strengthening existing self-supervised methods by explicitly enforcing a strong 3D structural prior directly into the model during training. Through experiments, across a range of datasets, we demonstrate that our 3D aware representations are more robust compared to conventional self-supervised baselines. 
\vspace{-8pt}
\end{abstract}

\section{Introduction}
\label{sec:intro}
The visual stimuli processed by a binocular, actively moving, human observer provides direct information about the 3D world around them~\cite{gibson1950perception}. 
As a result, humans have a remarkable ability to perceive useful 3D shape cues, enabling them to interact and navigate adeptly in complex environments. 
Most impressively, the power of the human visual system is not understood to be a resulting property of supervised learning, \ie it has developed thanks largely to `self-supervision'~\cite{smith2005development}. 

While great advances have been made in the past decade in developing computer vision systems, their success can be mostly attributed to large-scale \emph{supervised} representation learning. 
Moreover, current artificial vision systems are not yet nearly as robust as the human equivalent~\cite{geirhos2018imagenet}. 
For example, existing commonly used architectures are known to heavily rely on texture cues, which results in sub-optimal generalization performance~\cite{geirhos2018imagenet,naseer2021intriguing}. 
Encouragingly, neural networks that also make use of more shape cues have also been observed to be more robust to different types of image distortions~\cite{geirhos2018imagenet}.

\begin{figure}[t]
  \centering
  \begin{subfigure}[t]{0.32\columnwidth}
    \centering
    \includegraphics[width=\columnwidth]{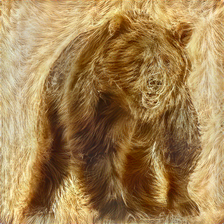}
    \caption{}
  \end{subfigure}
  \centering
  \begin{subfigure}[t]{0.32\columnwidth}
    \centering
    \includegraphics[width=\columnwidth]{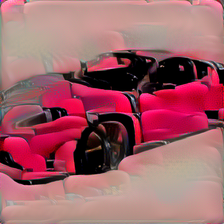}
    \caption{}
  \end{subfigure}
  \centering
  \begin{subfigure}[t]{0.32\columnwidth}
    \centering
    \includegraphics[width=\columnwidth]{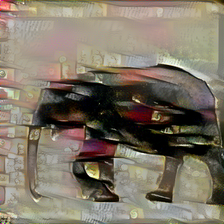}
    \caption{}
  \end{subfigure}
  \vspace{-5pt}
  \caption{Humans have no difficulty in recognizing the categories depicted in the above images, even though the texture of the objects has been perturbed. 
  This is thought to be in large part due to our reliance on shape, as opposed to texture, cues~\cite{landau1988importance,spelke2007core, geirhos2018imagenet}. However, an automated recognition system built on top of a state-of-the-art self-supervised representation learning approach (\ie DINOv2~\cite{oquab2023dinov2}) classifies these examples as dog, chair, and knife respectively, as the texture of the images resembles those object classes. 
  We introduce a new approach to improve the robustness of self-supervised methods using a proxy 3D reconstruction task which encourages representations that emphasize shape cues more. 
  As a result, our approach correctly predicted these examples as bear, car, and elephant. 
  }
  \label{fig:into_motivation}
  \vspace{-10pt} 
\end{figure}

These observations point to two important questions that are potentially hindering our artificial vision systems: (i) how do we reduce the over-reliance on supervised labeled data and (ii) how do we encourage models to make greater use of shape information to improve their robustness? 
Thankfully, great progress has been made on the first question as we now have methods for obtaining effective visual representations through self-supervision alone, \eg~\cite{wu2018unsupervised,chen2020simple,bao2021beit,he2022masked,oquab2023dinov2}. 
While methods exist for extracting shape-adjacent information in the form of depth using self-supervision from collections of image pairs~\cite{godard2017unsupervised} or video sequences~\cite{zhou2017unsupervised}, these approaches tend to require strong assumptions about the scenes they are trained on (\eg smooth camera motion, static scenes, limited visual diversity, \etc). 
As a result, the current most effective approaches for predicting depth require explicit depth supervision during training~\cite{Ranftl2022}. 
Moreover, even when depth supervision is available, it is not trivial to use it to improve the performance on other tasks~\cite{zamir2018taskonomy,standley2020tasks}.

In this work, we attempt to address these combined challenges by proposing a new method to improve existing self-supervised representation learning approaches by enforcing these models to reason about object/scene shape during training. 
We build on recent advances in 3D generative modeling~\cite{chan2022efficient,3dgp} to develop a self-supervised reconstruction method that generates a 3D representation of the input image. 
Our model is trained with a self-supervised reconstruction objective, starting from an already trained self-supervised network (\eg \cite{oquab2023dinov2}). 
Given an input image, we first extract a global feature representation using a pre-trained backbone network and then predict a 3D representation of the scene depicted in the image. 
Then we reconstruct appearance and depth maps using volume rendering from the predicted 3D representation. We use the difference between the reconstructed image and the original input image, and the difference between the predicted depth map and its pseudo ground truth as our training objectives. We do not utilize any manual labels during training as we only require an unordered (\ie not from videos or stereo pairs) collection of monocular images and their corresponding estimated depth from a previously trained depth prediction model~\cite{bhat2023zoedepth} as input. 
To minimize the training loss, the learned image representation needs to capture details about the shapes of the objects depicted in the input scenes.

While conceptually simple, the advantage of our approach is that it works with monocular image collections and does not make strong assumptions about the types of images it is trained on. 
As a result, we can train it using standard representation learning datasets such as ImageNet~\cite{russakovsky2015imagenet}. 
Quantitative and qualitative results illustrate that our shape-aware representations are more robust compared to variants that are not shape aware on a variety of downstream tasks. 
See Figure~\ref{fig:into_motivation} for a qualitative example. 

In summary, we make the following contributions: (i) We explore the role of 3D information when performing self-supervised learning on unordered monocular image collections. (ii) We propose a new method that enhances self-supervised learned representations via a proxy task that explicitly encodes 3D knowledge during training. (iii) When applied to a range of robustness tasks, our approach obtains superior performance compared to baselines that do not make use of 3D information at training time. 

\section{Related work}
\label{sec:rl}
In this section, we discuss related work in self-supervised learning, monocular shape understanding, and the role of shape in visual recognition.

\noindent\textbf{Self-supervised learning.} Recent approaches for deep learning-based self-supervised learning (SSL) in computer vision can be categorized into two groups: (i) predictive methods, where the learning objective depends solely on the input image, and (ii) discriminative approaches, which use additional images as inputs. 

Predictive tasks include context prediction (\eg patch or pixel prediction from a masked input image)~\cite{doersch2015unsupervised,chen2020generative,bao2021beit,zhou2021ibot,he2022masked}, colorization of grayscale input images~\cite{zhang2016colorful}, in-painting of randomly selected areas~\cite{pathak2016context}, predicting image rotation~\cite{gidaris2018unsupervised}, or object counting in the input image~\cite{noroozi2017representation}. 
In contrast, discriminative approaches aim to learn representations that make the input image, and an augmented version of it, more similar to each other compared to other randomly selected images~\cite{hadsell2006dimensionality,wu2018unsupervised,dosovitskiy2014discriminative,chen2020simple,oquab2023dinov2,oord2018representation,grill2020bootstrap, zbontar2021barlow}. 
Regularization to prevent trivial solutions~\cite{grill2020bootstrap,zbontar2021barlow,oquab2023dinov2} and selecting challenging negative examples~\cite{he2020momentum,chen2020improved} are important considerations for these methods. 
It is worth noting that some of the above methods make use of both types of losses. 
For a more comprehensive overview of SSL approaches, we direct the reader to~\cite{jing2020self,gui2023survey}. 

One limitation of the above approaches is that their focus is on 2D representation learning. 
In this work, we aim to enhance the robustness of self-supervised networks by utilizing a 3D proxy task during training. 
Recently, \cite{yu2023mvimgnet} introduced a new dataset consisting of common everyday objects containing multiple images, from different camera viewpoints, for each object instance. 
Their dataset is significantly larger than existing comparable multi-view datasets (\eg~\cite{henzler2021unsupervised}). 
They use this data to perform view-consistent self-supervised fine-tuning and show that this pseudo-3D supervision results in better downstream image classification performance on their dataset. 
However, multi-view data of this form is still very cumbersome and time-consuming to collect and thus current datasets are still limited in their scope. 

Recently, a new synthetic dataset named Photorealistic Unreal Graphics (PUG)~\cite{bordes2023pug} was introduced. 
It could be used as a source of multi-view data as the images are rendered using 3D assets. 
However, the images lack realism and the diversity of objects is still not on par with large scale 2D image collections. 
In this work, we show that it is possible to inject 3D information into a self-supervised model by training on single-view (\ie not multi-view) image collections alone.

\noindent\textbf{Single-view 3D understanding.} Our approach uses a proxy monocular 3D reconstruction task during training to enhance SSL performance. 
There is also a body of work that aims to estimate 3D shape from monocular images where their focus is on generation and not representation learning. 

Example existing works estimate partial 3D shape in the form of depth maps, \ie per-pixel continuous depth predictions. 
These methods either use pseudo ground truth depth supervision during training~\cite{Ranftl2022,bhat2023zoedepth} or are trained without depth supervision via image reconstruction losses~\cite{garg2016unsupervised,godard2017unsupervised,zhou2017unsupervised,godard2019digging}. 
Another line of work attempts to estimate the full 3D geometry of objects using 3D category priors using explicit representations like meshes~\cite{kanazawa2018end}, implicit representations like surface maps~\cite{guler2018densepose}, or with skeletons~\cite{wu2023magicpony}. 
The disadvantage of these methods is that they require strong category shape priors (\eg a 3D deformable model of a human). 
More recently, there have been some category-centric works that attempt to relax the need for strong shape priors~\cite{monnier2022share,aygun2023saor}. 
However, these are still category focused and are thus limited to specific classes of objects that have well-defined shapes (\eg animals or humans). 

The specific choice of 3D representation (\eg volume, mesh, or points) used by these methods can have a big impact on the quality of the 3D generated outputs and the computation required to train the model. 
In the last few years, implicit 3D representations parameterized via neural networks have become widely adopted for a range of 3D tasks~\cite{niemeyer2020differentiable,yu2021pixelnerf,mildenhall2021nerf}. 
However, conventional implicit networks can be very slow to train, which hinders their applicability to large-scale SSL. 
To address this, in this work, we make use of efficient implicit representations popularized by methods that perform 3D generative modeling from single image collections~\cite{chan2022efficient,3dgp}. 

\noindent\textbf{Shape and semantics.} Finally, we review work that utilizes shape information for visual recognition. 
It is well established, especially in the early years of cognitive development, that infants more heavily rely on shape cues compared to other cues such as texture during early category learning~\cite{landau1988importance,spelke1990principles,spelke2007core}. However, computational methods like CNNs~\cite{geirhos2018imagenet} and Vision Transformers~\cite{naseer2021intriguing,geirhos2021partial} do the opposite. 
With more data, and bigger models, there is some evidence to suggest that this over reliance on texture may decrease~\cite{dehghani2023scaling}, but it still does not fully disappear.

Prior to the wide adoption of deep-learning methods in computer vision, there were a large number of works that utilized (2D) shape information for recognition tasks. 
Examples include seminal works such as pictorial structures~\cite{fischler1973representation} and deformable templates~\cite{dalal2005histograms,felzenszwalb2010object,jain2011handbook,pepik20123d}. 
Subsequently, end-to-end trained approaches that did not use any structure or shape overtook these methods. %
However, recently a new set of methods have been developed that illustrate the benefit of using explicit shape information when combined with end-to-end learning methods for tasks like tracking~\cite{rajasegaran2022tracking} and action recognition~\cite{rajasegaran2023benefits}. 

Furthermore, recent studies have employed alternative forms of training data, such as styled images~\cite{geirhos2018imagenet} and edge maps~\cite{mummadi2021does}, to enhance shape awareness, albeit in a supervised context. In this work, we take inspiration from human cognition to add more shape information into our models by developing a proxy 3D reconstruction task to enhance SSL. 
To solve the resulting 3D reconstruction task, our model needs to learn more about the shape, and not just the texture, of objects during training.

\section{Method}
\label{sec:method}

The aim of visual representation learning is to learn a function that can map an input image into a representation. This is achieved by optimizing an objective function on a set of training data. For self-supervised learning (SSL), the objective function is optimized without using human-provided supervision~\cite{wu2018unsupervised,chen2020simple,bao2021beit,he2022masked,oquab2023dinov2}. 
However, given the lack of large-scale and semantically diverse datasets containing 3D information, current self-supervised methods are typically limited to using 2D unordered image collection during training. As a result, the learned representations that emerge from models trained on 2D images are not necessarily fully capable of capturing all the properties of the 3D world~\cite{geirhos2021partial}. In this work, we aim to improve these learned representations by using an additional proxy 3D task during training. Our aim is not to learn a new function from scratch, but to instead improve an existing pre-trained one.

\begin{figure*}[h]
  \centering
  \includegraphics[width=\textwidth]{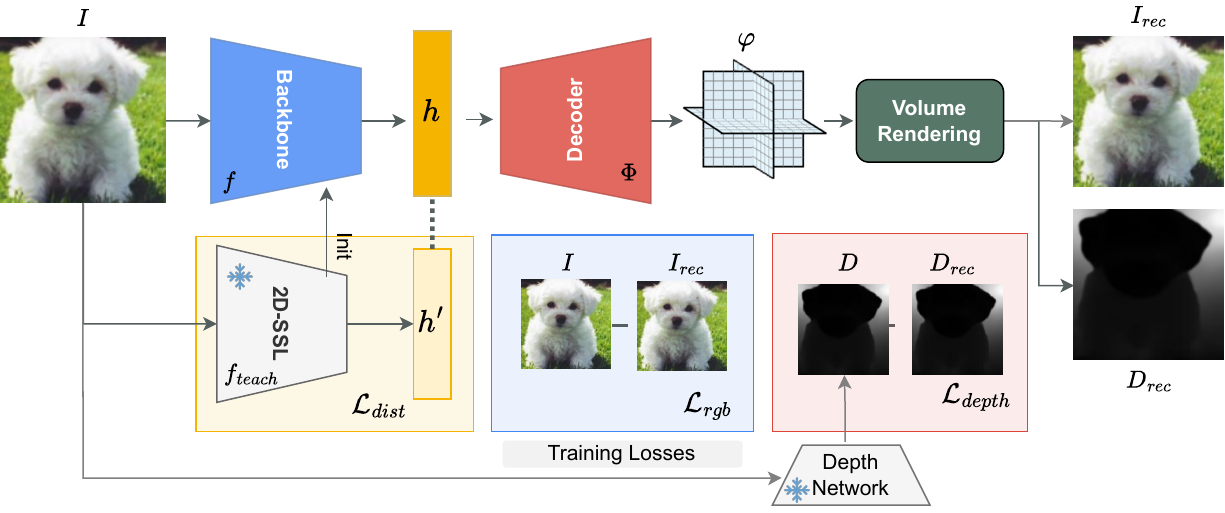} 
  \vspace{-12pt}
  \caption{ Overview of our self-supervised single-view 3D reconstruction approach. Given an input image, $I$, we first extract a representation of the image using an encoder network, $h = f(I)$. Then using a decoder network, $\Phi$, we generate triplane features~\cite{chan2022efficient,3dgp}. Using volume rendering~\cite{mildenhall2021nerf}, conditioned on a fixed camera location, we reconstruct the input image, $I_{rec}$, and its depth $D_{rec}$. We optimize all networks using a combination of reconstruction losses on the input image, $\mathcal{L}_{rgb}$, and estimated depth, $\mathcal{L}_{depth}$, along with a distillation loss, $\mathcal{L}_{dist}$, from a frozen 2D self-supervised learning model to prevent the forgetting of already learned informative representations. 
  }
  \label{fig:method_overview}
  \vspace{-8pt}
\end{figure*}

\subsection{Background}
Our goal is to learn a representation function $f(.)$, represented as a neural network, that can map an input image $I$ into a representation $h=f(I)$, where $h\in\mathcal{R}^d$. 
This will be achieved by optimizing an objective function $\mathcal{L}$ on a set of training data, without using any manually labeled data. We want to improve networks that are trained without supervision, since it has been shown in~\cite{goldblum2023battle} that SSL-based backbones like DINOv2~\cite{oquab2023dinov2} outperform supervised counterparts as a global image representation. 

There are a large number of publicly available self-supervised models that can extract useful representations from 2D images. Therefore, instead of learning a new representation function from scratch, we utilize a backbone that is already pre-trained with 2D self-supervised methods such as DINOv2~\cite{oquab2023dinov2}, and improve it by training it on a new proxy 3D task.

\subsection{3D aware robust representation learning }
We use 3D reconstruction as our proxy 3D task, \ie given an input monocular image, at training time, the network is tasked with reconstructing the 3D scene/objects depicted in the image. %
The intuition behind this is that for the network to successfully perform reconstruction, it must also learn 3D aware features from the input images.
As we want the network to learn image representations that transfer well to a large variety of scenes, the input images should be visually diverse and the 3D representation should be able to model complex scenes with multiple objects and diverse backgrounds. Moreover, the reconstruction task should be relatively computationally efficient to enable large-scale training. 
While there are alternative approaches for generating 3D predictions from 2D images~\cite{henzler2019escaping, pan20212d}, motivated by the need for efficiency we opted to use triplanes~\cite{chan2022efficient} as our 3D representation. 
Triplanes explicitly encode latent network features on axis-aligned planes. 
These features can then be aggregated via lightweight implicit feature decoders to perform efficient volume rendering for 3D reconstruction. 
Recently, \cite{3dgp} showed that triplanes can be used to generate 3D depictions of images for various types of scenes from visually diverse datasets such as ImageNet~\cite{russakovsky2015imagenet}.

Formally, given an input image $I$, we first extract a global image representation $h=f(I)$ using a backbone feature extractor network $f(.)$. 
This backbone can be pre-trained using a 2D self-supervised method. 
Then we use a decoder $\Phi(.)$ to generate triplane features from the representations of the input image. 
Note that we only require the decoder and triplane at training time and they can be discarded at inference as we only need to retain the backbone. 

This decoder takes the backbone features as input and produces triplane features $\varphi = \Phi(h), \varphi \in \mathcal{R}^{H\times W\times C\times 3}$. 
The decoder consists of two components, a set of learnable triplane embeddings $\xi\in \mathcal{R}^{(h\cdot w\cdot 3)\times D}$ and upsampling blocks. As done in~\cite{shen2023gina}, we first apply cross-attention between the triplane embeddings and the image representation to obtain low resolution triplane features, $\varphi' = \text{cross}(\xi, h), \varphi' \in \mathcal{R}^{h\times w\times C\times 3}$. Then we apply upsampling layers, which consist of bilinear upsampling and convolution operations, to obtain full resolution triplane features $\varphi = \text{upsample}(\varphi')$. 
Different than~\cite{shen2023gina}, we do not employ any quantization or style mapping~\cite{karras2020analyzing} as our goal is not to learn an unconditional generator, but to estimate 3D representation from the input image. 

To perform volume rendering, we compute the radiance field using a simple two-layer MLP similar to~\cite{3dgp,chan2022efficient} using features from the triplane at specified 3D points.
Note that as~\cite{chan2022efficient} and~\cite{3dgp} are generative methods, and not representation learning approaches, they generate triplanes from random codes. 
In contrast, our approach generates triplanes conditioned on the input image's representation $h$ which is obtained from the backbone network.
Using volume rendering from the triplane features, we produce the reconstructed image, $I_{rec}$, and its corresponding depth map, $D_{rec}$, 
\begin{align}
  I_{rec}, D_{rec} = \Pi(\Gamma(\varphi, \pi)),
\end{align}
where $\Gamma$ is the function that queries the radiance fields from triplanes conditioned on camera pose $\pi$ which contains extrinsic and intrinsic parameters, and $\Pi$ is volume a rendering operation~\cite{mildenhall2021nerf}. %
We use a fixed camera pose in our experiments since we want to learn a viewer-centered 3D representation, which is shown to be more generalizable compared to object-centric representations~\cite{shin2018pixels,tatarchenko2019single}. Moreover, as we reconstruct the whole scene with potentially multiple objects, the canonical pose is ambiguous. %

Given that 3D reconstruction from a 2D image is an ill-posed problem, like 3DGP~\cite{3dgp}, we make use of 2D depth information to produce plausible 3D predictions. 
As ground truth depth maps are not available for large-scale, in-the-wild, datasets like ImageNet~\cite{russakovsky2015imagenet}, we use pseudo ground truth depth maps obtained from off-the-shelf monocular depth methods such as ZoeDepth~\cite{bhat2023zoedepth}. Different from 3DGP~\cite{3dgp}, we do not modify the depth reconstruction with an adapter, but use it as it is for computing a depth reconstruction loss.

Given the input image $I^{i}$ and its pseudo depth $D^{i}$, we simply train the decoder such that it generates a plausible 3D prediction that is capable of reconstructing them both using the following losses 
\begin{align}
  \mathcal{L}_{rgb} & = \frac{1}{B}\sum_{i=1}^{B} ||I^{i} - I_{rec}^{i}||_{2},\\
  \mathcal{L}_{depth} & = \frac{1}{B}\sum_{i=1}^{B} ||D^{i} - D_{rec}^{i}||_{2},
\end{align} 
where B is the batch size.
Here, $\mathcal{L}_{rgb}$ and $\mathcal{L}_{depth}$ represent mean squared error losses, that depend on the input image and its depth. 
An illustration of our overall pipeline is depicted in Figure~\ref{fig:method_overview}. In addition to the reconstruction losses, we apply a L1 normalization loss for the density values the radiance fields, $\mathcal{L}_{norm} = \sum_{i=1}^{B} || \Delta^{i} ||_{1}$, where $\Delta^{i}$ is the set of density values that calculated for all the queried 3D points for the image $I^{i}$.

\subsection{Preventing forgetting} 
The benefit of our approach is that we can apply it to any self-supervised network that has already been pre-trained using 2D objectives. 
However, our 3D reconstruction objective might inadvertently bias the model towards the 3D task and force it to `forget' the useful representation that it has already encoded. 
To prevent this, we add a knowledge distillation loss~\cite{hinton2015distilling}, 
\begin{align}
  \mathcal{L}_{dist} = \frac{1}{B}\sum_{i=1}^{B} ||f(I^{i}) - f_{teach}(I^{i})||_{2}.
\end{align}
Here, $f(.)$ is the representation function that we are optimizing and $f_{teach}$ is a frozen backbone that is already trained using a 2D self-supervised objective. 

Our final overall training objective consists of a combination of four losses
\begin{align}
  \mathcal{L} = \lambda_{rgb}\mathcal{L}_{rgb} + \lambda_{depth}\mathcal{L}_{depth} + \lambda_{dist}\mathcal{L}_{dist} + \lambda_{norm}\mathcal{L}_{norm},
\end{align}
where the $\lambda$ values are weights for each of the respective loss terms. 

\subsection{Implementation details} 

\textbf{Backbone}: We implement our approach using different variants of standard Vision Transformers (ViTs)~\cite{dosovitskiy2020image}. Unless otherwise stated, for each experiment we start with a backbone network that has been pre-trained with the state-of-the-art SSL DINOv2~\cite{oquab2023dinov2} method. We first extract class and patch tokens from the last four layers of the backbone networks, and concatenate these features to obtain a global image representation which is of size $16 \times 16 \times D$, where D depends on the backbone architecture. 

\noindent\textbf{Decoder}: To generate the triplane features, we combine self-attention with learnable embeddings and 2D upsampling convolution layers similar to~\cite{shen2023gina}. We learn $16 \times 16 \times 3 \times D$ triplane embeddings, and by applying cross attention and upsampling blocks we obtain our final triplane features that are of size $128 \times 128 \times 3 \times D$.
Finally, we reconstruct the image and depth map using volume rendering~\cite{barron2021mip}, which results in an output resolution of $256 \times 256 \times 3$ and $256 \times 256$, for the reconstructed image and depth respectively. For volume rendering, for each pixel we first sample a ray that passes through pixel location and sample 16 points along the ray to obtain 3D points that we want to estimate the radiance field. Similar to previous works~\cite{3dgp,barron2021mip}, we utilize importance sampling. Then, we bilinearly sample the feature of each 3D points from the triplane, and calculate the radiance field (occupancy and rgb color) using a two layer MLP and pass these values to the final rendering operation to obtain the final reconstruction.

\noindent\textbf{Training}: For our training dataset, we use ImageNet-1k~\cite{russakovsky2015imagenet} which contains approximately 1.2 million images depicting 1,000 object classes.
We extract pseudo depth maps for all training images using ZoeDepth~\cite{bhat2023zoedepth} with the DPT backbone. 
We optimize all of the components of the reconstruction network (\eg backbone, decoder, and triplane embeddings) in an end-to-end manner using the Adam~\cite{kingma2014adam} optimizer for 10 epochs with a fixed learning rate of $1e^{-4}$. We set $\lambda_{rgb} = 0.1, \lambda_{depth} = 1, \lambda_{dist} = 1$ and $\lambda_{norm} = 1e^{-3}$ in all of our experiments. 
During training, we only use basic random crops and horizontal flip augmentations. 

\section{Experiments}
\label{sec:exps}

The main goal of our proposed method is to enhance the robustness of existing representation learning methods. 
We first show how our method results in improved performance on several robustness benchmarks such as ImageNet-Rendition (Im-R)~\cite{hendrycks2021many}, ImageNet-Sketch (Im-Sketch)~\cite{wang2019learning}, and Photorealistic Unreal Graphics (PUG)~\cite{bordes2023pug}. %
We also perform experiments on conventional tasks like image recognition~\cite{russakovsky2015imagenet}, fine-grained image classification~\cite{van2021benchmarking}, and depth estimation~\cite{silberman2012indoor}. 
This is to illustrate that our approach does not decrease performance for other tasks at the expense of improving robustness. 

After training the network with the proxy 3D task, we discard the 3D estimation components of the network and use the backbone representation function to extract global image representation from images $h = f(I)$. 
For evaluation, we train per-task decoder networks using a fixed representation function $y = \psi(h)$, where the form of $y$ and $\psi(h)$ depends on the specific downstream task. 
In each experimental section, we provide details about the decoding function and training details. In all of our experiments, the representation functions are frozen, unless stated otherwise.

\subsection{Robustness}

\textbf{Datasets.} We present experimental results on benchmarks that are designed to test the robustness of methods in the face of various appearance related shifts. 
ImageNet-Rendition (Im-R)~\cite{hendrycks2021many} contains 30,000 images of art, cartoon, graffiti, \etc from 200 ImageNet classes. 
ImageNet-Sketch (Im-Sketch)~\cite{wang2019learning} contains 50 sketch images for each of the 1000 original ImageNet classes. 
These two datasets contain examples where the texture of the objects is significantly different compared to real in-the-wild photographs, which leads to a significant drop in performance for previous SSL methods~\cite{oquab2023dinov2}.

Photorealistic Unreal Graphics (PUG)~\cite{bordes2023pug} is a dataset that is designed to evaluate the robustness of visual recognition models. 
It contains synthetically generated examples from 3D assets by controlling for factors like object texture, background, lighting \etc. 
It has been demonstrated that state-of-the-art visual recognition models obtain inadequate performance on this dataset due to changes in appearance factors like object texture and size~\cite{bordes2023pug}. 
We report performance on these two main factors in our experiments. 

The 3D Common Corruptions (ImageNet-3DCC)~\cite{kar20223d} dataset was created with synthetic corruptions with varying levels of difficulty using ImageNet validation images. 
Compared to previous datasets, it was constructed with synthetic augmentations, but it contains real images and realistic corruptions such as low lighting, flash, and motion blur which reflect real-world challenges for visual recognition models.

\noindent\textbf{Protocol.} All of the experiments that we present here are designed to measure the robustness of classifiers that are trained on ImageNet~\cite{russakovsky2015imagenet} classification data. Given this, we first train a linear classifier on top of various frozen backbones from DINOv2~\cite{oquab2023dinov2} that are either enhanced via our method (denoted as `+ 3D-Prior') or not, using 1k ImageNet classes from the original training set. 
We then test the respective linear classifiers on various robustness datasets. 

\noindent\textbf{Results.} In Table~\ref{tab:robustness}, we observe that our proposed method (`3D-Prior') improves the robustness of SSL methods on all robustness benchmarks tested, irrespective of architecture type. 
For instance, we improve the performance for the different backbone architectures on both ImageNet-Rendition and ImageNet-Sketch datasets, which contain highly challenging out-of-distribution examples. 
In particular, the performance of DINOv2~\cite{oquab2023dinov2} using the ViTB/14 architecture is improved by 2\% on both benchmarks. 
Furthermore, for the PUG benchmark, our method improved the performance of the models for object size and texture variation in all cases. 

We also present results on ImageNet-3DCC dataset for various synthetic corruption types in Table~\ref{tab:imagenet_3dcc}. For each level, there is 5 different corruption levels, for simplicity we report averaged the top-1 accuracy for each corruption type. We observe slight improvement for corruption types like far focus, xy motion blur and z motion blurs. However, for other factors like low light, iso noise, we achieve performance that is comparable to the baselines.

\begin{figure*}
  \centering
  \includegraphics[height=1.695cm, trim=10 0 10 0, clip]{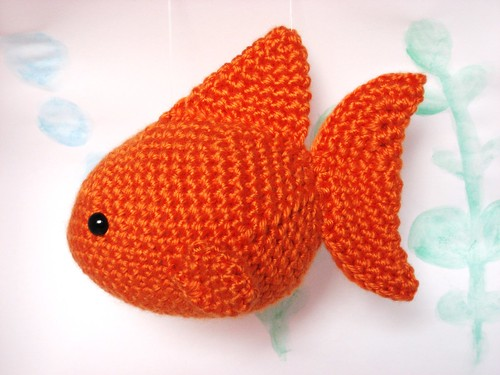} 
  \includegraphics[height=1.695cm, trim=10 50 10 50, clip]{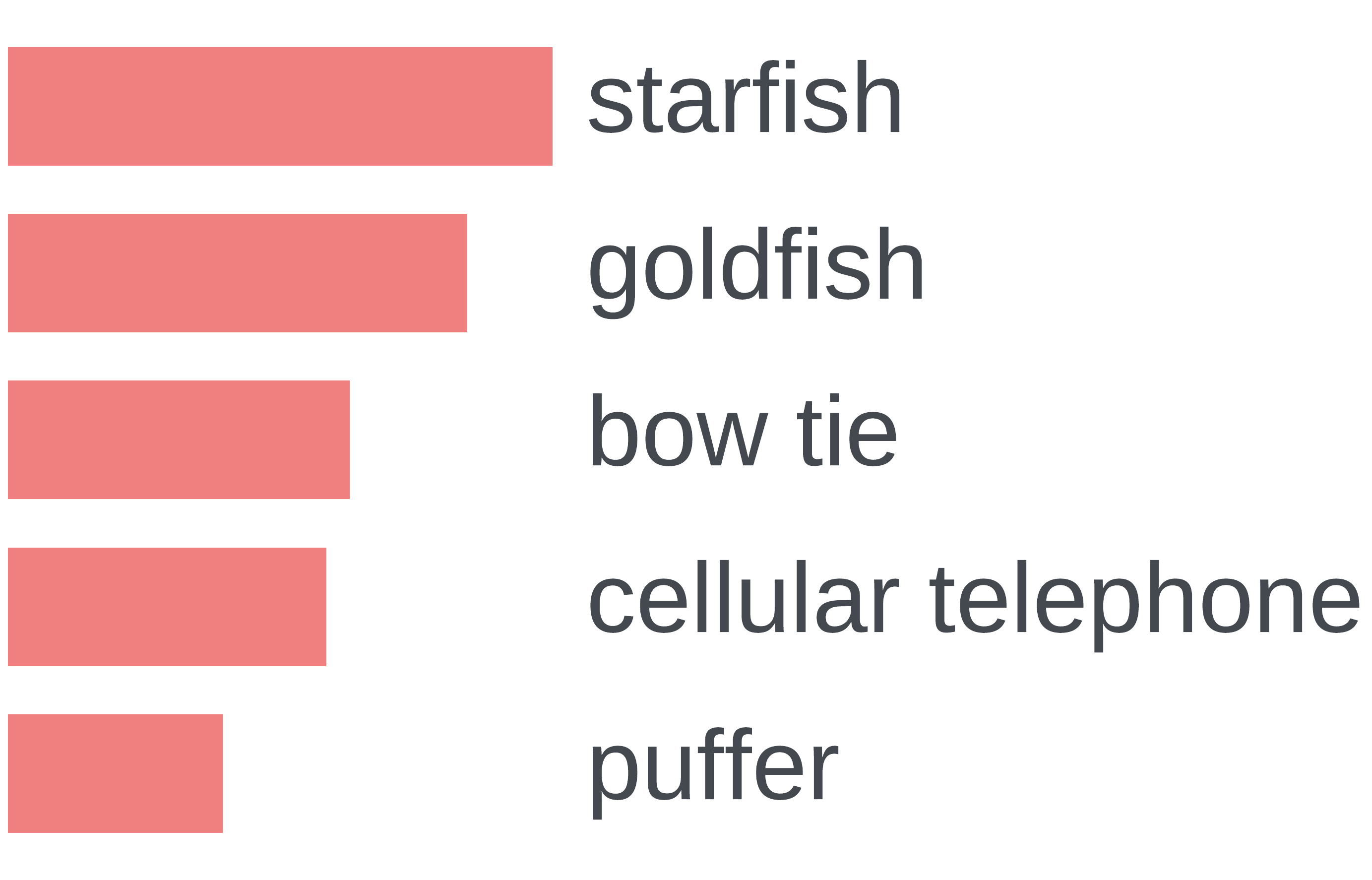} 
  \includegraphics[height=1.695cm, trim=10 50 10 50, clip]{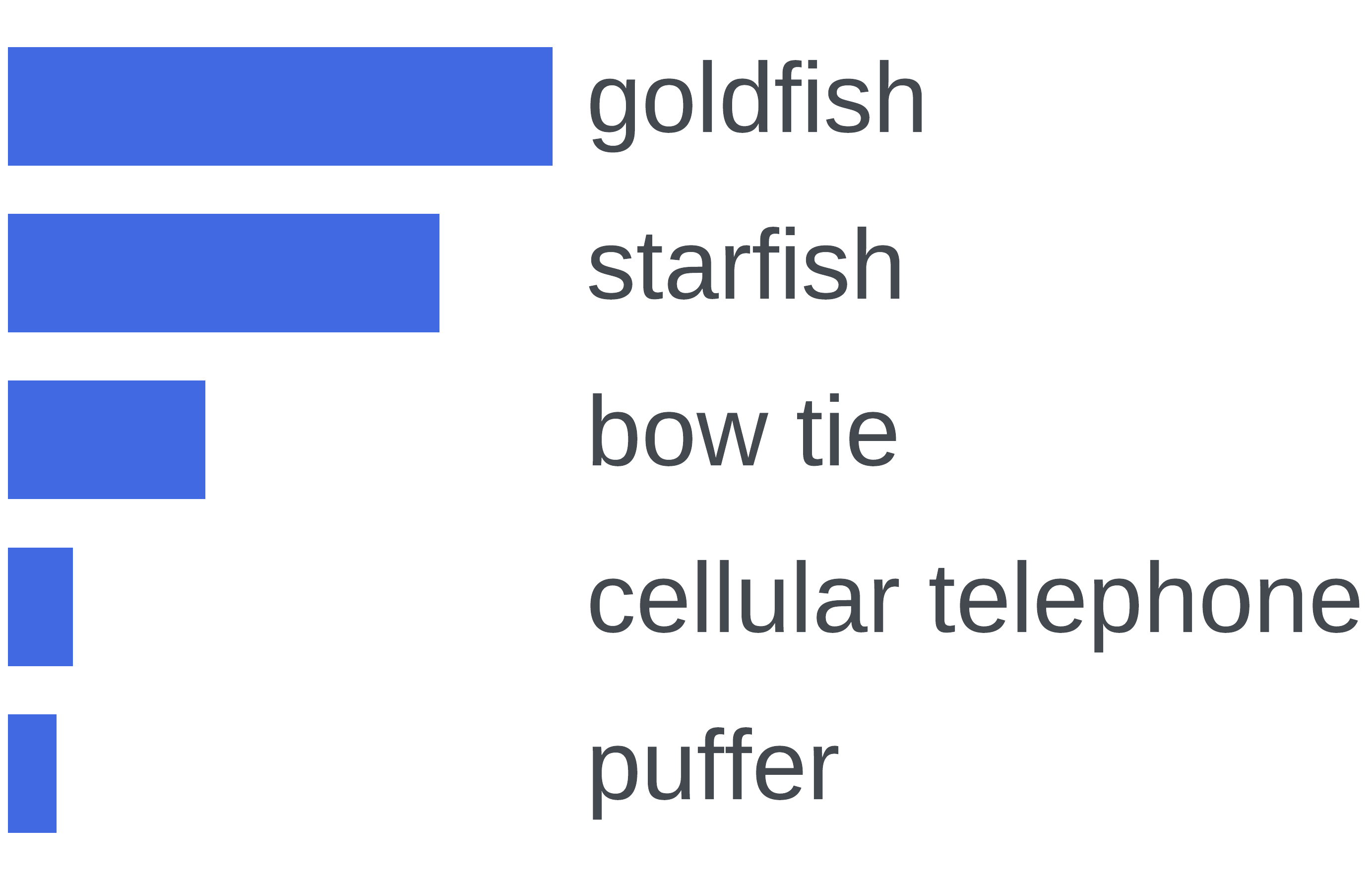}
  \centering
  \quad\quad\quad
  \includegraphics[height=1.695cm, trim=0 0 0 0, clip]{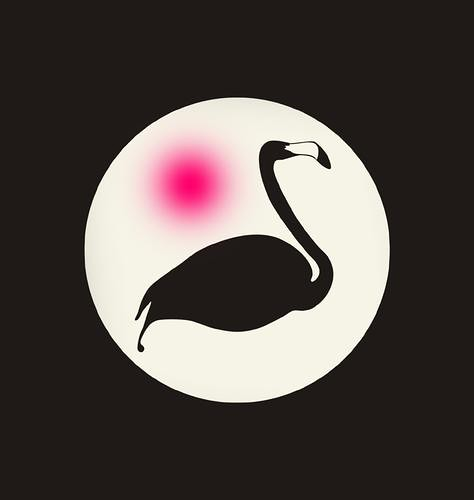} 
  \includegraphics[height=1.695cm, trim=10 50 10 50, clip]{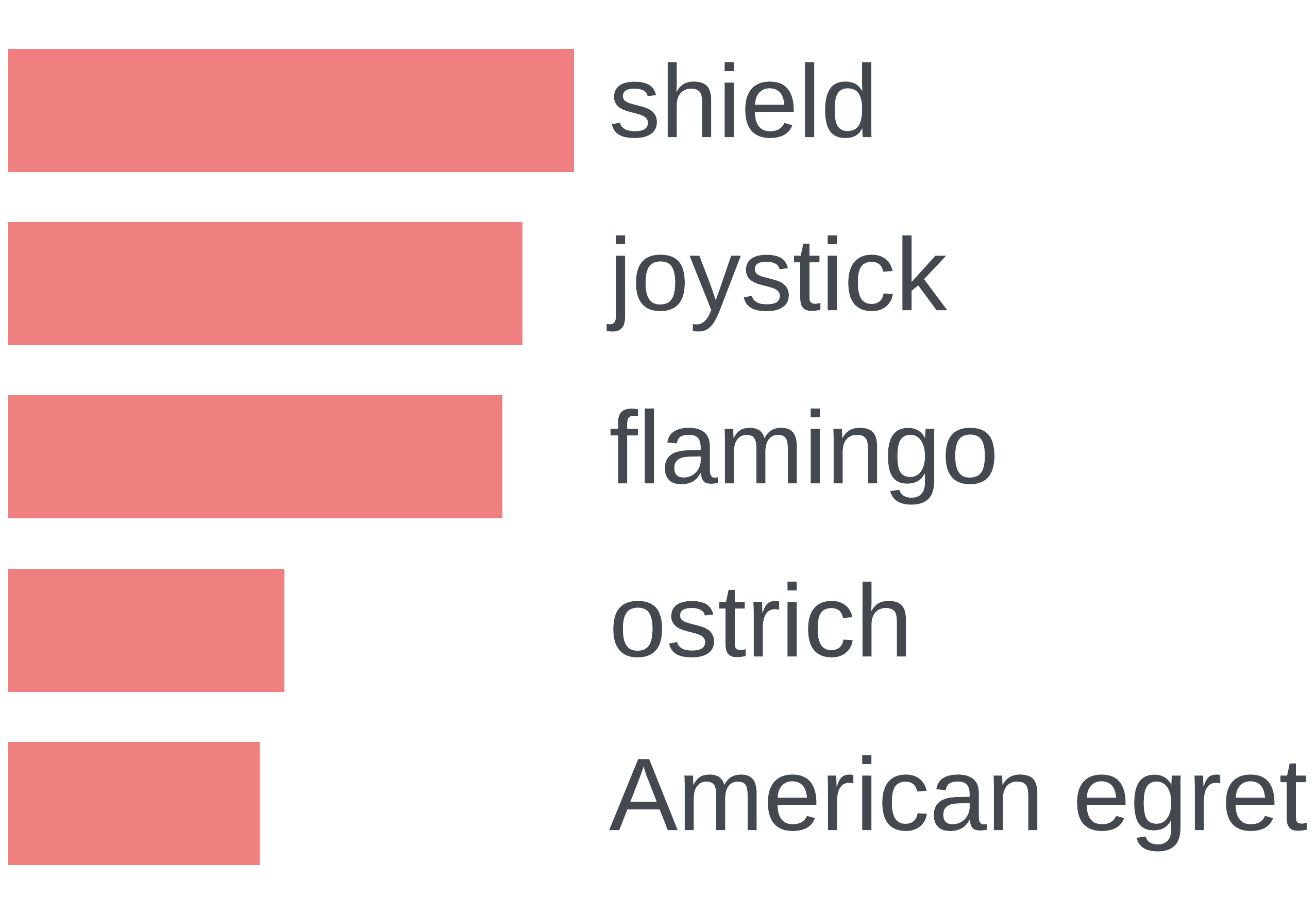} 
  \includegraphics[height=1.695cm, trim=10 50 10 50, clip]{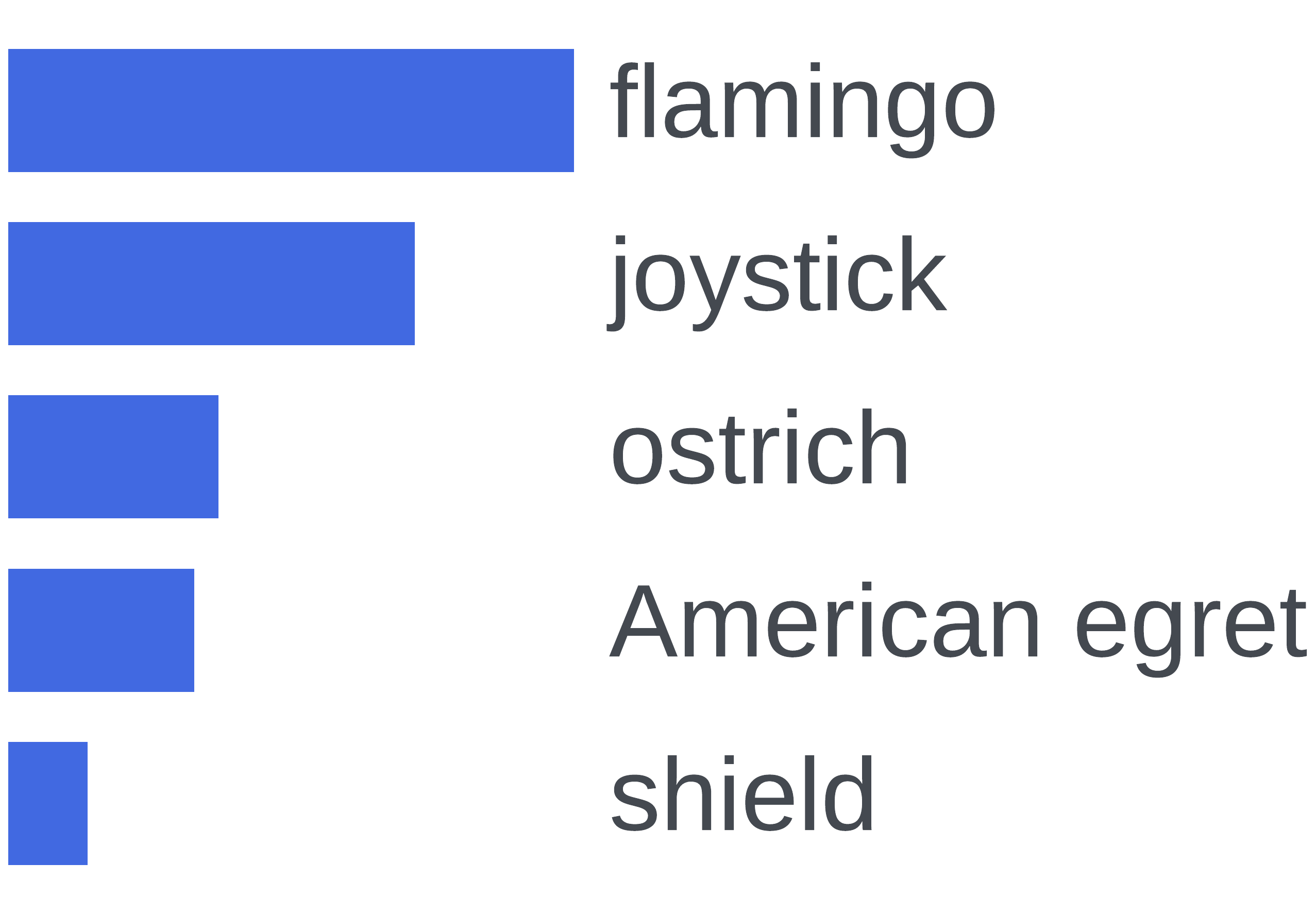} 
  
  \vspace{0.5cm}
  
  \centering
  \includegraphics[height=1.70cm, trim=0 15 0 10, clip]{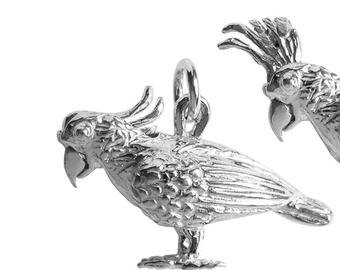} 
  \includegraphics[height=1.70cm, trim=0 0 0 0, clip]{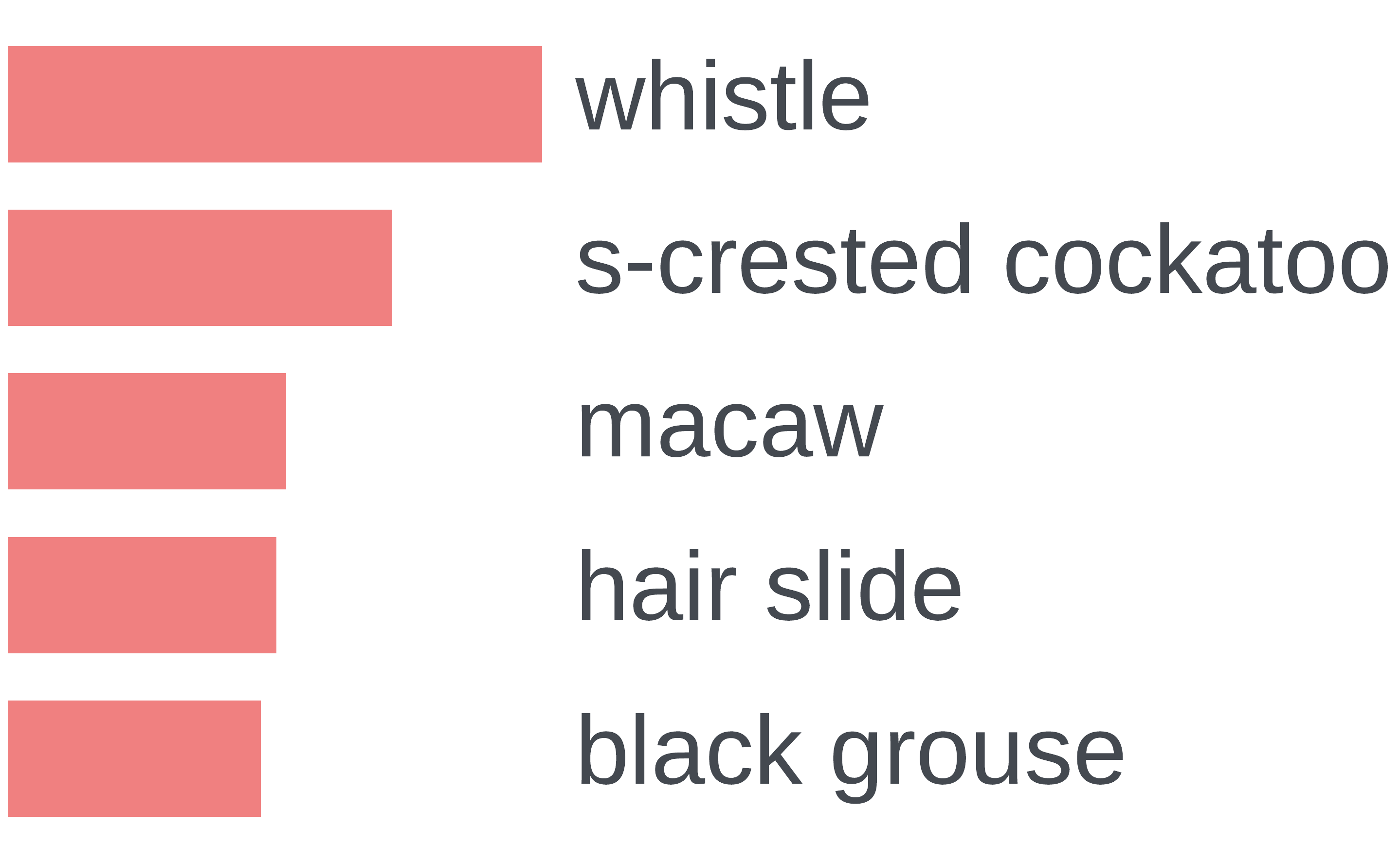} 
  \includegraphics[height=1.70cm, trim=0 0 0 0, clip]{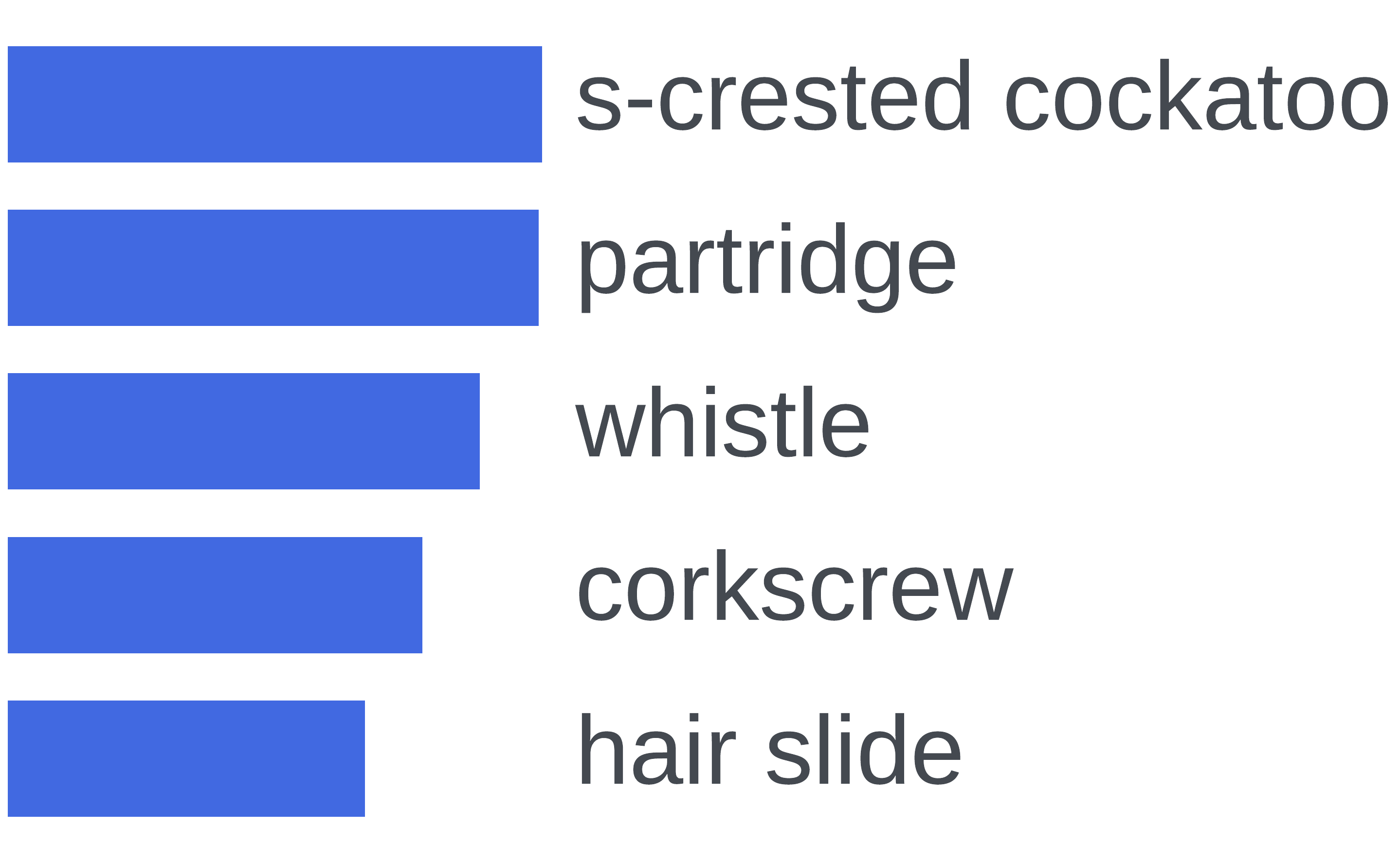}
  \quad\quad\quad
  \includegraphics[height=1.68cm, trim=0 50 0 0, clip]{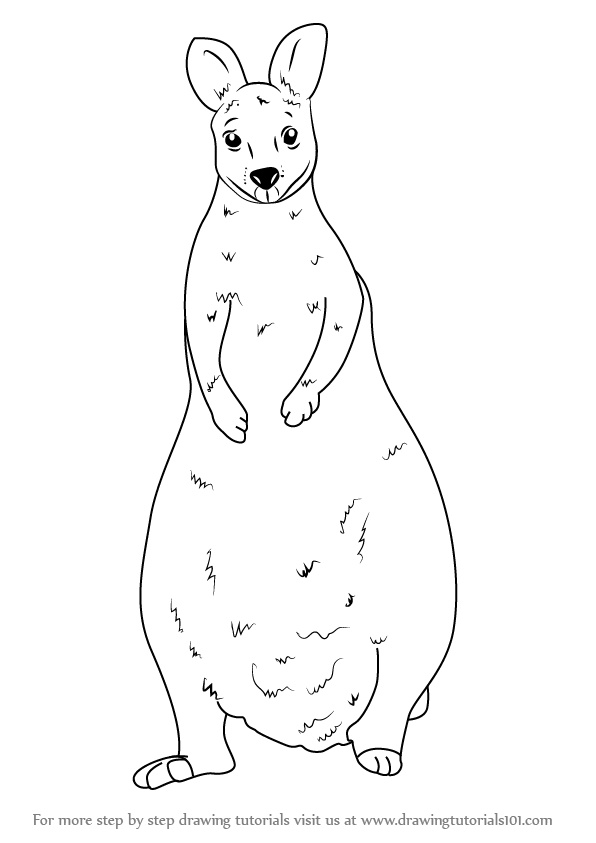} 
  \quad
  \includegraphics[height=1.68cm, trim=10 50 10 50, clip]{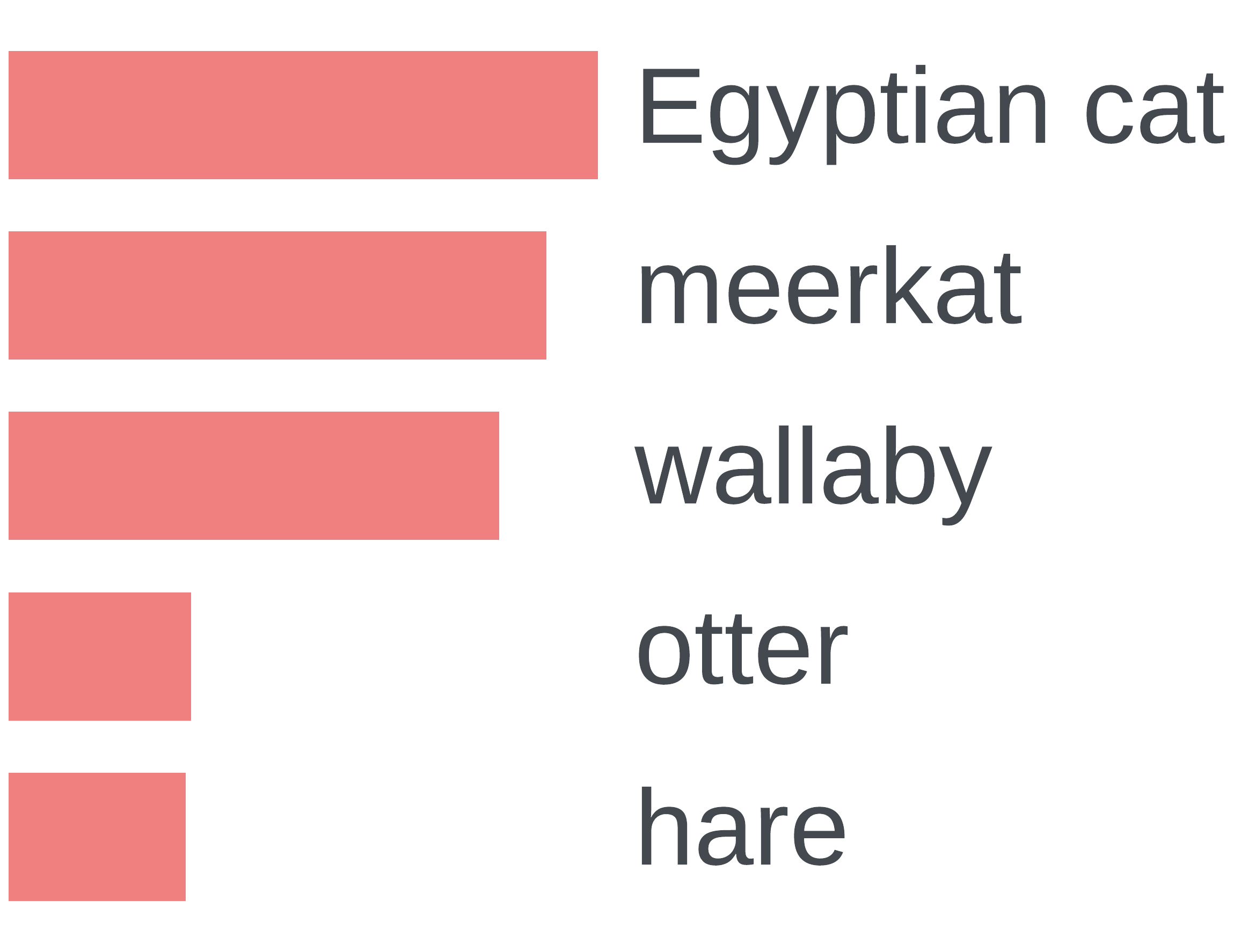} 
  \quad\quad
  \includegraphics[height=1.68cm, trim=10 50 10 50, clip]{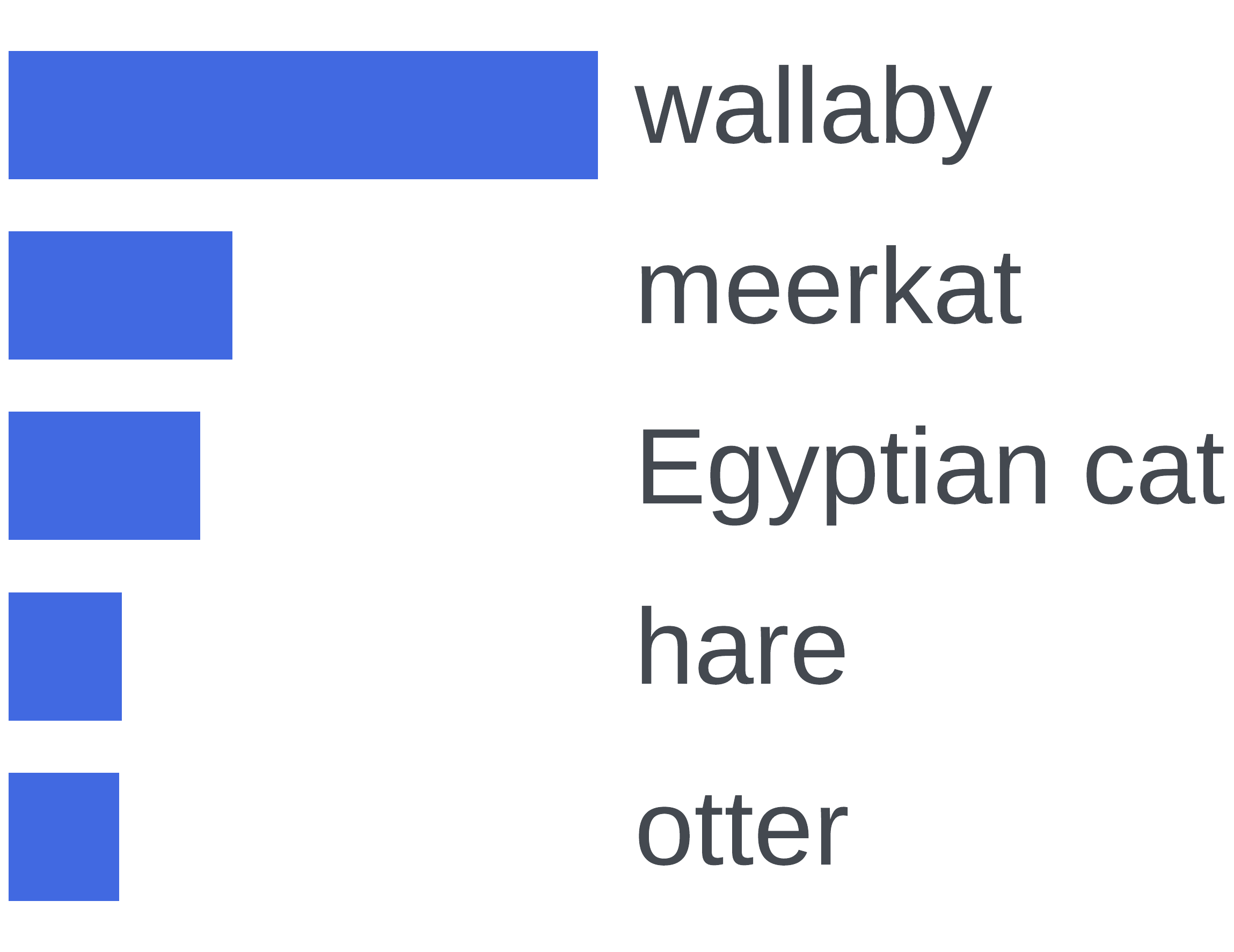}
  
  \caption{Here we compare top-5 predictions from linear classifiers that are trained on original DINOv2~\cite{oquab2023dinov2} backbone features (shown in red) and our 3D enhanced approach (shown in blue) on various challenging examples from ImageNet-Rendition~\cite{hendrycks2021many} and ImageNet-Sketch~\cite{wang2019learning}. 
  Our method results in more shape information being encoded in the representation and hence leads to classifiers that are more robust for these challenging out-of-distribution examples.
  }
  \vspace{-12pt}
  \label{fig:results1}
\end{figure*}

We also present qualitative results in Figure~\ref{fig:results1}. 
We illustrate some top-5 predictions from linear classifiers that were trained on top of representations from DINOv2, either with or without our method. 
For instance, the top left example is misclassified as a `starfish' by the DINOv2-based classifier due to the color of the input image while our \emph{shape-aware} approach correctly identifies the images as containing a `goldfish' due to improved shape-bias.

\begin{table}
 \centering
 \resizebox{0.99\columnwidth}{!}{
 \begin{tabular}{l c c c c}
 \toprule
   Method & Im-R & Im-Sketch & PUG-Texture & PUG-Size\\ \midrule
   ViT-S/14 & 53.7 & 41.2 & 20.7 & 26.8 \\
   ViT-S/14 + 3D-Prior & \textbf{54.6} & \textbf{41.8} & \textbf{21.2} & \textbf{26.9} \\ \midrule
   ViT-B/14 & 63.3 & 50.6 & 25.3 & 32.2 \\
   ViT-B/14 + 3D-Prior & \textbf{65.9} & \textbf{52.4} & \textbf{26.2} & \textbf{33.4} \\ \midrule
   ViT-L/14 & 74.4 & 59.3 & 34.5 & 42.7 \\
   ViT-L/14 + 3D-Prior & \textbf{75.9} & \textbf{59.5} & \textbf{36.4} & \textbf{43.2}\\ 
   \bottomrule
 \end{tabular}
 }
 \vspace{-5pt}
 \caption{Robustness evaluation using frozen backbone features from DINOv2~\cite{oquab2023dinov2} and their enhanced versions from our method (`+ 3D-Prior'). Here we report top-1 accuracy for all benchmarks. 
 Irrespective of backbone architecture type, our 3D-Prior method improves performance on across all datasets. 
 For the PUG experiments we re-run the DINOv2 baselines with our evaluation setting.
 }
 \vspace{-10pt}
 \label{tab:robustness}
\end{table}

\begin{table*}
 \centering
 \resizebox{0.99\linewidth}{!}{
 \begin{tabular}{l c c c c c c c c c}
 \toprule
   Method & color quant & far focus & flash & fog 3d & iso noise & low light & near focus & xy motion blur & z motion blur
\\ \midrule
  ViT-B/14 & 72.5 & 71.6 & \textbf{60.8} & 62.5 & \textbf{63.9} & 72.3 & 75.5 & 58.5 & 58.3 \\ 
  ViT-B/14 + 3D-Prior & \textbf{72.6} & \textbf{72.0} & 60.7 & \textbf{62.7} & 63.3 & 72.3 & \textbf{75.8} & \textbf{59.0} & \textbf{58.7} \\
   \bottomrule
 \end{tabular}
 }
 \vspace{-5pt}
 \caption{Robustness evaluation using frozen backbone features from DINOv2~\cite{oquab2023dinov2} and their enhanced versions from our method on the ImageNet-3DCC dataset~\cite{kar20223d} using a ViT-B/14-based architecture. While our method improves robustness for corruptions such as `motion blur' and `far focus', there are cases such as `flash' where we are slightly worse.}
 \label{tab:imagenet_3dcc}
\end{table*}

\subsection{Downstream tasks}

\textbf{Tasks and datasets.} We present results on additional downstream tasks to show that our method does not lead to worse performance for other tasks at the expense of improved robustness. 
We report results for visual recognition on ImageNet~\cite{russakovsky2015imagenet}, fine-grained classification using the iNaturalist 2021~\cite{van2021benchmarking} and depth estimation on NYU-DepthV2~\cite{silberman2012indoor}.

\noindent\textbf{Protocol.} We follow the same evaluation protocol as in DINOv2~\cite{oquab2023dinov2}. 
For ImageNet and iNat21 experiments we froze the backbone, and trained a single-layer classifier using the respective training sets and reported the top-1 validation accuracy. 
For depth estimation on NYU-DepthV2, we trained two different decoders on top of frozen backbone features, a single linear layer and a more complex DPT~\cite{Ranftl2022} decoder and followed the same training recipe from DINOv2~\cite{oquab2023dinov2}. 
We also compare to the non-3D baseline numbers from~\cite{oquab2023dinov2}.

\noindent\textbf{Results.} Results are presented in Table~\ref{tab:downstream}. For the visual recognition task on ImageNet-1k, we observe that linear classification performance is slightly improved for all the backbone architectures evaluated and the performance of the models on fine-grained classification for iNat21 is maintained. 
Furthermore, the performance on depth estimation is improved compared to the baselines, especially when we use a high-resolution DPT decoder on top of our learned representation.

\begin{table}
 \centering
 \resizebox{0.99\columnwidth}{!}{
 \begin{tabular}{l c c c c}
 \toprule
   \multirow{2}{*}{Method} & \multirow{2}{*}{ImageNet-1k} & \multirow{2}{*}{iNat21} & \multicolumn{2}{c}{NYU-DepthV2 $\downarrow$} \\ \cmidrule{4-5}
   & & & linear & DPT \\ \midrule
   ViT-S/14 & 81.1 & \textbf{74.2} & 0.499& 0.356 \\
   ViT-S/14 + 3D-Prior & \textbf{81.4} & 73.6 & \textbf{0.438}& \textbf{0.346} \\ \midrule
   ViT-B/14 & 84.5 & 81.1 & 0.399& 0.317 \\
   ViT-B/14 + 3D-Prior & \textbf{85.1} & \textbf{82.0} & \textbf{0.398}& \textbf{0.300} \\ \midrule
   ViT-L/14 & 86.3 & 85.1 & \textbf{0.384 }& 0.293 \\
   ViT-L/14 + 3D-Prior & \textbf{86.5} & \textbf{85.2} & 0.389 & \textbf{0.286} \\ 
   \bottomrule
 \end{tabular}
 }
 \vspace{-2pt}
 \caption{Downstream task evaluation using frozen backbone features on various tasks using DINOv2~\cite{oquab2023dinov2} with and without our 3D-Prior method. We report top-1 accuracy for the ImageNet-1k~\cite{russakovsky2015imagenet} and iNat21~\cite{van2021benchmarking} datasets (higher is better), and RMSE for NYU-DepthV2~\cite{silberman2012indoor} dataset (lower is better). 
 Our method leads to improvements in visual recognition performance on ImageNet and for depth estimation on NYU-DepthV2, and does not negatively impact performance on the fine-grained iNat21 dataset. 
 }
 \vspace{-8pt}
 \label{tab:downstream}
\end{table}

\subsection{Shape bias}

\begin{figure}[t]
  \centering
  \includegraphics[width=0.9\columnwidth]{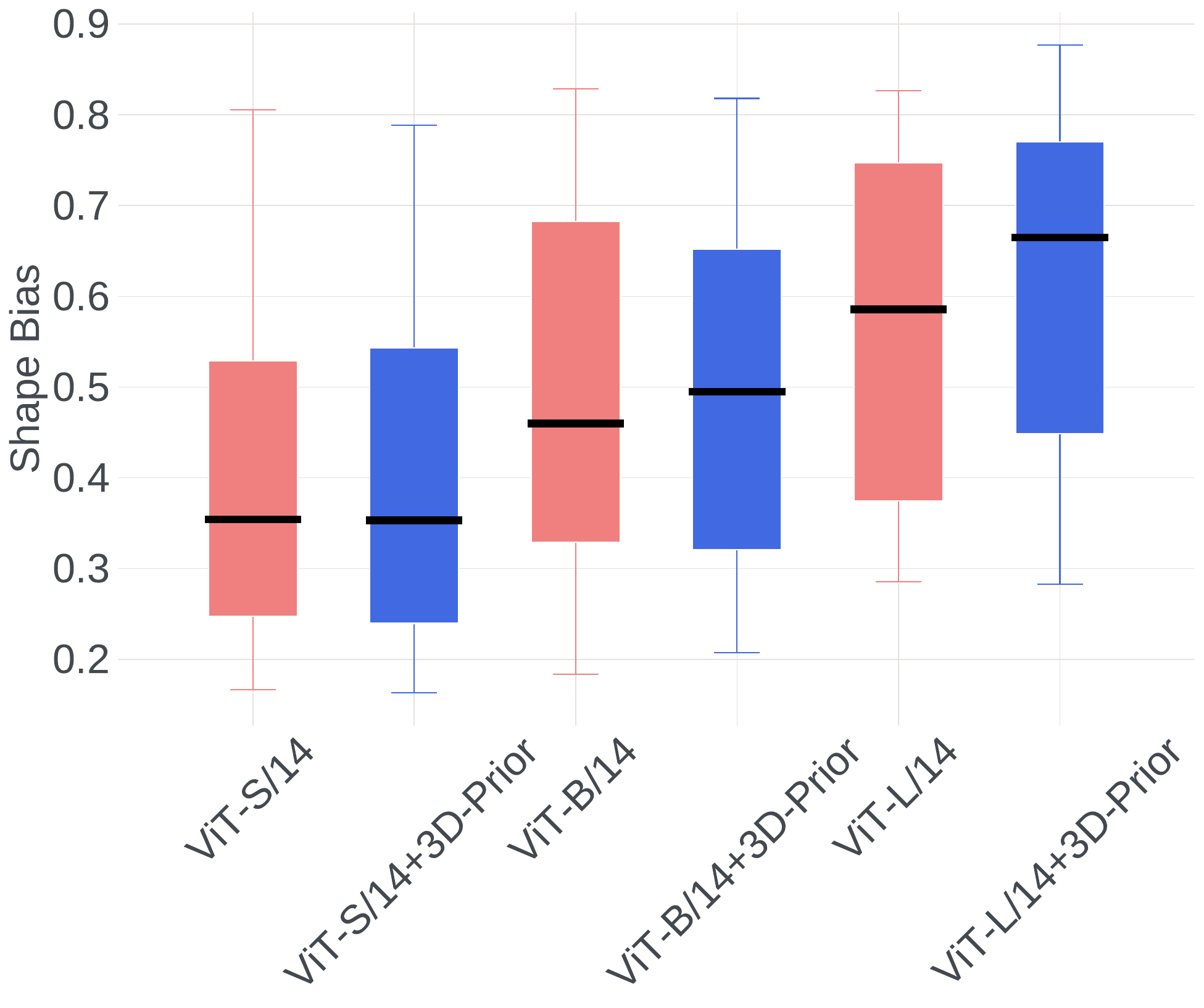} 
  \vspace{-5pt}
  \caption{Quantification of the shape bias of different DINOv2~\cite{oquab2023dinov2} representations with and without our 3D-Prior method. We calculate the shape bias using the data and protocol from~\cite{geirhos2018imagenet}. 
  Our approach increases the shape bias of visual recognition models and we observe that with larger backbones, the difference grows.
  }
  \vspace{-10pt}
  \label{fig:shape_bias}
\end{figure}

Similar to humans, we want our visual recognition models to pay more attention to shape cues compared to texture. 
As our proxy 3D task requires learning more shape-oriented representations, our hypothesis is that it should lead to representations that have more shape bias. 
We use the same experimental protocol and dataset from~\cite{geirhos2018imagenet} to measure the shape bias of different models. 
The dataset contains various synthetically generated examples, where the shape of the object comes from one class and the texture of the object comes from another. 

We measure the shape bias of representations from DINOv2~\cite{oquab2023dinov2} before and after it is trained with our proxy 3D objective. 
The results are visualized in Figure~\ref{fig:shape_bias}. 
We observe that our method improves the shape bias of the original representations, which is the objective of our shape-centric 3D reconstruction task. 
Qualitative examples, where we compare predictions of models with and without our method, can be seen in Figure~\ref{fig:into_motivation}. 

These results, combined with the robustness experiments, show that the hypothesis of improving shape bias to obtain more robust representations is valid. 
Furthermore, these results may further encourage future lines of work in SSL to develop methods that are designed to explicitly consider 3D representations during training.

\subsection{Ablations}

\begin{table}
 \centering
 \resizebox{0.99\columnwidth}{!}{
 \begin{tabular}{l c c c c c}
 \toprule
   Method & Im-1k & Im-R & Im-Sketch & PUG-Texture & PUG-Size\\ \midrule
   DinoV2 & 84.5 & 63.3 & 50.6 & 25.3 & 32.2 \\ \midrule
   Ours & 85.1 & 65.9 & 52.4 & 26.2 & 33.4 \\ 
   Ours w/o triplane & 84.3 & 63.1 & 50.7 & 24.9 & 31.1 \\ 
   Ours w/o $\mathcal{L}_{dist}$ & 70.7 & 34.2 & 22.9 & 12.1 & 16.4 \\ 
   Ours from scratch & 14.4 & 8.5 & 7.4 & 0.2 & 0.1\\ 
   \bottomrule
 \end{tabular}
 }
 \vspace{-5pt}
 \caption{Ablation experiments on various robustness benchmarks using a DinoV2 ViT-B/14 model. We investigate the importance of using an explicit 3D representation during training (\ie w/o triplane), disabling our distillation loss (\ie w/o $\mathcal{L}_{dist}$), and we evaluate if we can learn reasonable representations when training from scratch without using a pre-trained backbone or distillation loss (\ie from scratch).
 }
 \label{tab:ablation}
 \vspace{-6pt}
\end{table}

To quantify the importance of individual components of our model, we present ablation experiments on the robustness tasks in Table~\ref{tab:ablation}.

\noindent{\bf Removing the triplane.} First we investigate if using an 3D representation in the form of a triplane with volume rendering is necessary or if training a basic depth and image decoder network on top of representations is sufficient. 
Here, we added a decoder which consists of multiple upsampling and convolution layers to predict depth and images. %
We observe a drop in performance on all benchmarks, but with a smaller drop on ImageNet. 
This experiment indicates that using an explicit 3D representation is crucial to improve the robustness of the learned representations.

\noindent{\bf Removing distillation.} Next, we try to understand what happens if we do not employ a distillation loss. 
Without distillation, the model is free to forget useful representation that are already encoded in the 2D SSL backbone. 
To test this, we simply trained a separate model without the distillation loss. 
Removing distillation leads to a significant drop in performance across all benchmarks. 
This experiment shows that preventing the forgetting of already learned useful representations is essential. 

\noindent{\bf Training from scratch.} We also investigate if we can learn a global image representation using only the proxy 3D task. Here, we initialized the backbone network randomly (\ie they are no longer pre-trained) and trained the network using only the image and depth reconstruction losses. The experimental results show that the learned representation is not meaningful and, by itself, the 3D reconstruction task is not a sufficient way to learn a global image representation.

\begin{figure}[t]
  \centering
  \includegraphics[width=0.99\columnwidth]{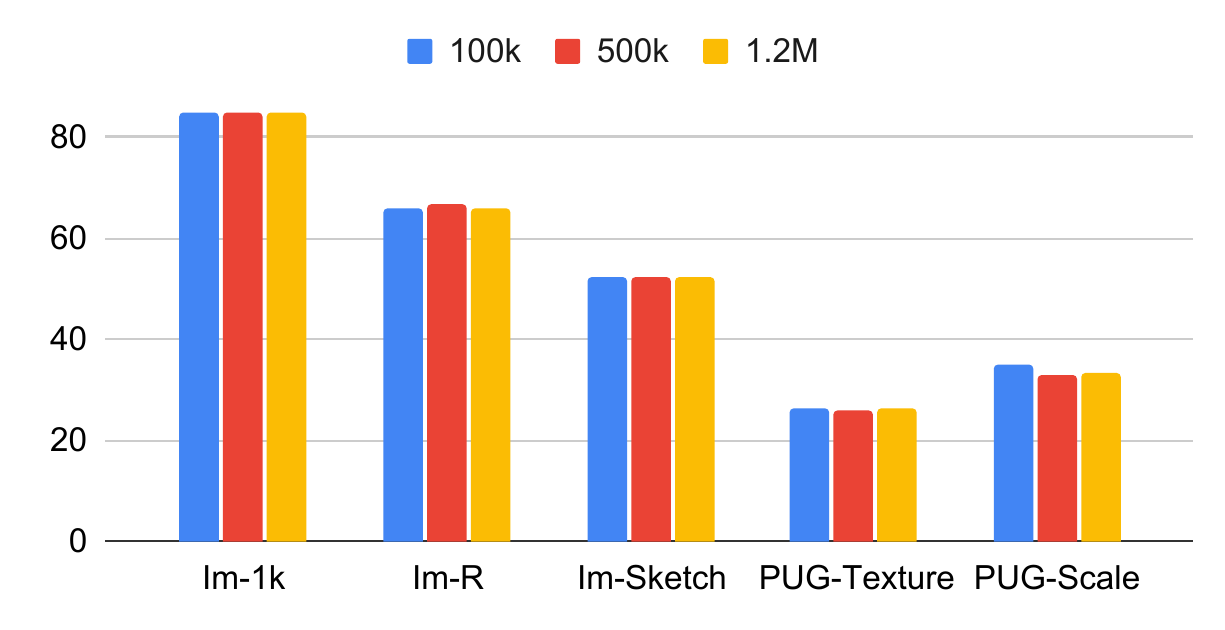} 
  \vspace{-10pt}
  \caption{We compare the performance of our approach using different amounts of training data from that same source for the 3D proxy task with DinoV2 ViT-B/14 backbones. Surprisingly, we observe that more data does not change the performance drastically, which shows that our method is data efficient.
  }
  \vspace{-10pt}
  \label{fig:dataset_size}
\end{figure}

\noindent{\bf Amount of data.} Finally, we explore the impact of varying the size of the training dataset that is used for the 3D proxy task. 
\cite{cole2022does} showed that 2D-based SSL methods benefit from being trained on larger unlabeled datasets, but that there are diminishing returns after a certain amount for the methods they tested. 
Similarly, we quantify how efficient our method is in terms of the training data size. 
For this experiment, we randomly selected 100k and 500k images from the ImageNet training set, and trained different instances of our model on these subsets using the same number of iterations as the full model. 
We report the results in Figure~\ref{fig:dataset_size}. 
Interestingly, compared to 2D self-supervised methods~\cite{cole2022does}, the performance of our 3D enhanced models are not significantly impacted by the reduction in training data. 

\subsection{Limitations}

One of the restrictions of our approach is the requirement for pseudo depth maps for each input image during training. 
Existing pre-trained monocular depth estimation models are used, \eg~\cite{Ranftl2022,bhat2023zoedepth}, and these pre-trained depth models only provide depth supervision and do not provide any semantic signal. 
Hence, these depth maps are free and easy to obtain for large-scale 2D image collections in an automatic way. 
More importantly, these types of depth estimation models have been demonstrated to be robust and generalizable to various settings. %

\section{Conclusion}
\label{sec:conclusion}

We presented a new approach to enhance the robustness of visual representations from 2D self-supervised networks. 
Our method utilizes a conceptually simple single-view 3D reconstruction task to encourage learning more shape-aware 3D centric representations. 
One of the distinct advantages of our approach is that it can be applied to unordered single image collections as it does not impose any stringent assumptions on the types of images it is trained on. 
We show that incorporating shape-aware knowledge into the representation learning process enhances robustness when compared to alternatives that are not shape aware across a range of visual understanding benchmarks. 
We hope that our results will encourage a new line of self-supervised works that are designed to consider 3D representations during training.

\clearpage
{
  \small
  \bibliographystyle{ieeenat_fullname}
  \bibliography{main}

\begin{thebibliography}{77}
\providecommand{\natexlab}[1]{#1}
\providecommand{\url}[1]{\texttt{#1}}
\expandafter\ifx\csname urlstyle\endcsname\relax
  \providecommand{\doi}[1]{doi: #1}\else
  \providecommand{\doi}{doi: \begingroup \urlstyle{rm}\Url}\fi

\bibitem[Ayg{\"u}n and Mac~Aodha(2024)]{aygun2023saor}
Mehmet Ayg{\"u}n and Oisin Mac~Aodha.
\newblock {SAOR: Single-View Articulated Object Reconstruction}.
\newblock In \emph{CVPR}, 2024.

\bibitem[Bao et~al.(2022)Bao, Dong, Piao, and Wei]{bao2021beit}
Hangbo Bao, Li Dong, Songhao Piao, and Furu Wei.
\newblock Beit: Bert pre-training of image transformers.
\newblock In \emph{ICLR}, 2022.

\bibitem[Barron et~al.(2021)Barron, Mildenhall, Tancik, Hedman, Martin-Brualla,
  and Srinivasan]{barron2021mip}
Jonathan~T Barron, Ben Mildenhall, Matthew Tancik, Peter Hedman, Ricardo
  Martin-Brualla, and Pratul~P Srinivasan.
\newblock Mip-nerf: A multiscale representation for anti-aliasing neural
  radiance fields.
\newblock In \emph{ICCV}, 2021.

\bibitem[Bhat et~al.(2023)Bhat, Birkl, Wofk, Wonka, and
  M{\"u}ller]{bhat2023zoedepth}
Shariq~Farooq Bhat, Reiner Birkl, Diana Wofk, Peter Wonka, and Matthias
  M{\"u}ller.
\newblock Zoedepth: Zero-shot transfer by combining relative and metric depth.
\newblock \emph{arXiv:2302.12288}, 2023.

\bibitem[Bordes et~al.(2023)Bordes, Shekhar, Ibrahim, Bouchacourt, Vincent, and
  Morcos]{bordes2023pug}
Florian Bordes, Shashank Shekhar, Mark Ibrahim, Diane Bouchacourt, Pascal
  Vincent, and Ari~S Morcos.
\newblock Pug: Photorealistic and semantically controllable synthetic data for
  representation learning.
\newblock \emph{arXiv:2308.03977}, 2023.

\bibitem[Chan et~al.(2022)Chan, Lin, Chan, Nagano, Pan, De~Mello, Gallo,
  Guibas, Tremblay, Khamis, et~al.]{chan2022efficient}
Eric~R Chan, Connor~Z Lin, Matthew~A Chan, Koki Nagano, Boxiao Pan, Shalini
  De~Mello, Orazio Gallo, Leonidas~J Guibas, Jonathan Tremblay, Sameh Khamis,
  et~al.
\newblock Efficient geometry-aware 3d generative adversarial networks.
\newblock In \emph{CVPR}, 2022.

\bibitem[Chen et~al.(2020{\natexlab{a}})Chen, Radford, Child, Wu, Jun, Luan,
  and Sutskever]{chen2020generative}
Mark Chen, Alec Radford, Rewon Child, Jeffrey Wu, Heewoo Jun, David Luan, and
  Ilya Sutskever.
\newblock Generative pretraining from pixels.
\newblock In \emph{ICML}, 2020{\natexlab{a}}.

\bibitem[Chen et~al.(2020{\natexlab{b}})Chen, Kornblith, Norouzi, and
  Hinton]{chen2020simple}
Ting Chen, Simon Kornblith, Mohammad Norouzi, and Geoffrey Hinton.
\newblock A simple framework for contrastive learning of visual
  representations.
\newblock In \emph{ICML}, 2020{\natexlab{b}}.

\bibitem[Chen et~al.(2020{\natexlab{c}})Chen, Fan, Girshick, and
  He]{chen2020improved}
Xinlei Chen, Haoqi Fan, Ross Girshick, and Kaiming He.
\newblock Improved baselines with momentum contrastive learning.
\newblock \emph{arXiv:2003.04297}, 2020{\natexlab{c}}.

\bibitem[Cole et~al.(2022)Cole, Yang, Wilber, Mac~Aodha, and
  Belongie]{cole2022does}
Elijah Cole, Xuan Yang, Kimberly Wilber, Oisin Mac~Aodha, and Serge Belongie.
\newblock When does contrastive visual representation learning work?
\newblock In \emph{CVPR}, 2022.

\bibitem[Dalal and Triggs(2005)]{dalal2005histograms}
Navneet Dalal and Bill Triggs.
\newblock Histograms of oriented gradients for human detection.
\newblock In \emph{CVPR}, 2005.

\bibitem[Dehghani et~al.(2023)Dehghani, Djolonga, Mustafa, Padlewski, Heek,
  Gilmer, Steiner, Caron, Geirhos, Alabdulmohsin, et~al.]{dehghani2023scaling}
Mostafa Dehghani, Josip Djolonga, Basil Mustafa, Piotr Padlewski, Jonathan
  Heek, Justin Gilmer, Andreas~Peter Steiner, Mathilde Caron, Robert Geirhos,
  Ibrahim Alabdulmohsin, et~al.
\newblock Scaling vision transformers to 22 billion parameters.
\newblock In \emph{ICML}, 2023.

\bibitem[Doersch et~al.(2015)Doersch, Gupta, and
  Efros]{doersch2015unsupervised}
Carl Doersch, Abhinav Gupta, and Alexei~A Efros.
\newblock Unsupervised visual representation learning by context prediction.
\newblock In \emph{ICCV}, 2015.

\bibitem[Dosovitskiy et~al.(2014)Dosovitskiy, Springenberg, Riedmiller, and
  Brox]{dosovitskiy2014discriminative}
Alexey Dosovitskiy, Jost~Tobias Springenberg, Martin Riedmiller, and Thomas
  Brox.
\newblock Discriminative unsupervised feature learning with convolutional
  neural networks.
\newblock \emph{NeurIPS}, 2014.

\bibitem[Dosovitskiy et~al.(2021)Dosovitskiy, Beyer, Kolesnikov, Weissenborn,
  Zhai, Unterthiner, Dehghani, Minderer, Heigold, Gelly,
  et~al.]{dosovitskiy2020image}
Alexey Dosovitskiy, Lucas Beyer, Alexander Kolesnikov, Dirk Weissenborn,
  Xiaohua Zhai, Thomas Unterthiner, Mostafa Dehghani, Matthias Minderer, Georg
  Heigold, Sylvain Gelly, et~al.
\newblock An image is worth 16x16 words: Transformers for image recognition at
  scale.
\newblock In \emph{ICLR}, 2021.

\bibitem[Felzenszwalb et~al.(2010)Felzenszwalb, Girshick, McAllester, and
  Ramanan]{felzenszwalb2010object}
Pedro~F Felzenszwalb, Ross~B Girshick, David McAllester, and Deva Ramanan.
\newblock Object detection with discriminatively trained part-based models.
\newblock \emph{PAMI}, 2010.

\bibitem[Fischler and Elschlager(1973)]{fischler1973representation}
Martin~A Fischler and Robert~A Elschlager.
\newblock The representation and matching of pictorial structures.
\newblock \emph{Transactions on Computers}, 1973.

\bibitem[Garg et~al.(2016)Garg, Bg, Carneiro, and Reid]{garg2016unsupervised}
Ravi Garg, Vijay~Kumar Bg, Gustavo Carneiro, and Ian Reid.
\newblock Unsupervised cnn for single view depth estimation: Geometry to the
  rescue.
\newblock In \emph{ECCV}, 2016.

\bibitem[Geirhos et~al.(2019)Geirhos, Rubisch, Michaelis, Bethge, Wichmann, and
  Brendel]{geirhos2018imagenet}
Robert Geirhos, Patricia Rubisch, Claudio Michaelis, Matthias Bethge, Felix~A
  Wichmann, and Wieland Brendel.
\newblock Imagenet-trained cnns are biased towards texture; increasing shape
  bias improves accuracy and robustness.
\newblock \emph{ICLR}, 2019.

\bibitem[Geirhos et~al.(2021)Geirhos, Narayanappa, Mitzkus, Thieringer, Bethge,
  Wichmann, and Brendel]{geirhos2021partial}
Robert Geirhos, Kantharaju Narayanappa, Benjamin Mitzkus, Tizian Thieringer,
  Matthias Bethge, Felix Wichmann, and Wieland Brendel.
\newblock Partial success in closing the gap between human and machine vision.
\newblock \emph{NeurIPS}, 2021.

\bibitem[Gibson(1950)]{gibson1950perception}
James~J Gibson.
\newblock The perception of the visual world.
\newblock 1950.

\bibitem[Gidaris et~al.(2018)Gidaris, Singh, and
  Komodakis]{gidaris2018unsupervised}
Spyros Gidaris, Praveer Singh, and Nikos Komodakis.
\newblock Unsupervised representation learning by predicting image rotations.
\newblock In \emph{ICLR}, 2018.

\bibitem[Godard et~al.(2017)Godard, Mac~Aodha, and
  Brostow]{godard2017unsupervised}
Cl{\'e}ment Godard, Oisin Mac~Aodha, and Gabriel~J Brostow.
\newblock Unsupervised monocular depth estimation with left-right consistency.
\newblock In \emph{CVPR}, 2017.

\bibitem[Godard et~al.(2019)Godard, Mac~Aodha, Firman, and
  Brostow]{godard2019digging}
Cl{\'e}ment Godard, Oisin Mac~Aodha, Michael Firman, and Gabriel~J Brostow.
\newblock Digging into self-supervised monocular depth estimation.
\newblock In \emph{ICCV}, 2019.

\bibitem[Goldblum et~al.(2023)Goldblum, Souri, Ni, Shu, Prabhu, Somepalli,
  Chattopadhyay, Ibrahim, Bardes, Hoffman, et~al.]{goldblum2023battle}
Micah Goldblum, Hossein Souri, Renkun Ni, Manli Shu, Viraj Prabhu, Gowthami
  Somepalli, Prithvijit Chattopadhyay, Mark Ibrahim, Adrien Bardes, Judy
  Hoffman, et~al.
\newblock Battle of the backbones: A large-scale comparison of pretrained
  models across computer vision tasks.
\newblock \emph{NeurIPS}, 2023.

\bibitem[Grill et~al.(2020)Grill, Strub, Altch{\'e}, Tallec, Richemond,
  Buchatskaya, Doersch, Avila~Pires, Guo, Gheshlaghi~Azar,
  et~al.]{grill2020bootstrap}
Jean-Bastien Grill, Florian Strub, Florent Altch{\'e}, Corentin Tallec, Pierre
  Richemond, Elena Buchatskaya, Carl Doersch, Bernardo Avila~Pires, Zhaohan
  Guo, Mohammad Gheshlaghi~Azar, et~al.
\newblock Bootstrap your own latent-a new approach to self-supervised learning.
\newblock \emph{NeurIPS}, 2020.

\bibitem[Gui et~al.(2023)Gui, Chen, Cao, Sun, Luo, and Tao]{gui2023survey}
Jie Gui, Tuo Chen, Qiong Cao, Zhenan Sun, Hao Luo, and Dacheng Tao.
\newblock A survey of self-supervised learning from multiple perspectives:
  Algorithms, theory, applications and future trends.
\newblock \emph{arXiv:2301.05712}, 2023.

\bibitem[G{\"u}ler et~al.(2018)G{\"u}ler, Neverova, and
  Kokkinos]{guler2018densepose}
R{\i}za~Alp G{\"u}ler, Natalia Neverova, and Iasonas Kokkinos.
\newblock Densepose: Dense human pose estimation in the wild.
\newblock In \emph{CVPR}, 2018.

\bibitem[Hadsell et~al.(2006)Hadsell, Chopra, and
  LeCun]{hadsell2006dimensionality}
Raia Hadsell, Sumit Chopra, and Yann LeCun.
\newblock Dimensionality reduction by learning an invariant mapping.
\newblock In \emph{CVPR}, 2006.

\bibitem[He et~al.(2020)He, Fan, Wu, Xie, and Girshick]{he2020momentum}
Kaiming He, Haoqi Fan, Yuxin Wu, Saining Xie, and Ross Girshick.
\newblock Momentum contrast for unsupervised visual representation learning.
\newblock In \emph{CVPR}, 2020.

\bibitem[He et~al.(2022)He, Chen, Xie, Li, Doll{\'a}r, and
  Girshick]{he2022masked}
Kaiming He, Xinlei Chen, Saining Xie, Yanghao Li, Piotr Doll{\'a}r, and Ross
  Girshick.
\newblock Masked autoencoders are scalable vision learners.
\newblock In \emph{CVPR}, 2022.

\bibitem[Hendrycks et~al.(2021)Hendrycks, Basart, Mu, Kadavath, Wang, Dorundo,
  Desai, Zhu, Parajuli, Guo, et~al.]{hendrycks2021many}
Dan Hendrycks, Steven Basart, Norman Mu, Saurav Kadavath, Frank Wang, Evan
  Dorundo, Rahul Desai, Tyler Zhu, Samyak Parajuli, Mike Guo, et~al.
\newblock The many faces of robustness: A critical analysis of
  out-of-distribution generalization.
\newblock In \emph{ICCV}, 2021.

\bibitem[Henzler et~al.(2019)Henzler, Mitra, and Ritschel]{henzler2019escaping}
Philipp Henzler, Niloy~J Mitra, and Tobias Ritschel.
\newblock Escaping plato's cave: 3d shape from adversarial rendering.
\newblock In \emph{ICCV}, 2019.

\bibitem[Henzler et~al.(2021)Henzler, Reizenstein, Labatut, Shapovalov,
  Ritschel, Vedaldi, and Novotny]{henzler2021unsupervised}
Philipp Henzler, Jeremy Reizenstein, Patrick Labatut, Roman Shapovalov, Tobias
  Ritschel, Andrea Vedaldi, and David Novotny.
\newblock Unsupervised learning of 3d object categories from videos in the
  wild.
\newblock In \emph{CVPR}, 2021.

\bibitem[Hinton et~al.(2015)Hinton, Vinyals, and Dean]{hinton2015distilling}
Geoffrey Hinton, Oriol Vinyals, and Jeff Dean.
\newblock Distilling the knowledge in a neural network.
\newblock \emph{arXiv:1503.02531}, 2015.

\bibitem[Jain and Li(2011)]{jain2011handbook}
Anil~K Jain and Stan~Z Li.
\newblock \emph{Handbook of face recognition}.
\newblock 2011.

\bibitem[Jing and Tian(2020)]{jing2020self}
Longlong Jing and Yingli Tian.
\newblock Self-supervised visual feature learning with deep neural networks: A
  survey.
\newblock \emph{PAMI}, 2020.

\bibitem[Kanazawa et~al.(2018)Kanazawa, Black, Jacobs, and
  Malik]{kanazawa2018end}
Angjoo Kanazawa, Michael~J Black, David~W Jacobs, and Jitendra Malik.
\newblock End-to-end recovery of human shape and pose.
\newblock In \emph{CVPR}, 2018.

\bibitem[Kar et~al.(2022)Kar, Yeo, Atanov, and Zamir]{kar20223d}
O{\u{g}}uzhan~Fatih Kar, Teresa Yeo, Andrei Atanov, and Amir Zamir.
\newblock 3d common corruptions and data augmentation.
\newblock In \emph{CVPR}, 2022.

\bibitem[Karras et~al.(2020)Karras, Laine, Aittala, Hellsten, Lehtinen, and
  Aila]{karras2020analyzing}
Tero Karras, Samuli Laine, Miika Aittala, Janne Hellsten, Jaakko Lehtinen, and
  Timo Aila.
\newblock Analyzing and improving the image quality of stylegan.
\newblock In \emph{CVPR}, 2020.

\bibitem[Kingma and Ba(2015)]{kingma2014adam}
Diederik~P Kingma and Jimmy Ba.
\newblock Adam: A method for stochastic optimization.
\newblock In \emph{ICLR}, 2015.

\bibitem[Landau et~al.(1988)Landau, Smith, and Jones]{landau1988importance}
Barbara Landau, Linda~B Smith, and Susan~S Jones.
\newblock The importance of shape in early lexical learning.
\newblock \emph{Cognitive development}, 1988.

\bibitem[Mildenhall et~al.(2021)Mildenhall, Srinivasan, Tancik, Barron,
  Ramamoorthi, and Ng]{mildenhall2021nerf}
Ben Mildenhall, Pratul~P Srinivasan, Matthew Tancik, Jonathan~T Barron, Ravi
  Ramamoorthi, and Ren Ng.
\newblock Nerf: Representing scenes as neural radiance fields for view
  synthesis.
\newblock \emph{Communications of the ACM}, 2021.

\bibitem[Monnier et~al.(2022)Monnier, Fisher, Efros, and
  Aubry]{monnier2022share}
Tom Monnier, Matthew Fisher, Alexei~A Efros, and Mathieu Aubry.
\newblock Share with thy neighbors: Single-view reconstruction by
  cross-instance consistency.
\newblock In \emph{ECCV}, 2022.

\bibitem[Mummadi et~al.(2021)Mummadi, Subramaniam, Hutmacher, Vitay, Fischer,
  and Metzen]{mummadi2021does}
Chaithanya~Kumar Mummadi, Ranjitha Subramaniam, Robin Hutmacher, Julien Vitay,
  Volker Fischer, and Jan~Hendrik Metzen.
\newblock Does enhanced shape bias improve neural network robustness to common
  corruptions?
\newblock In \emph{ICLR}, 2021.

\bibitem[Naseer et~al.(2021)Naseer, Ranasinghe, Khan, Hayat, Shahbaz~Khan, and
  Yang]{naseer2021intriguing}
Muhammad~Muzammal Naseer, Kanchana Ranasinghe, Salman~H Khan, Munawar Hayat,
  Fahad Shahbaz~Khan, and Ming-Hsuan Yang.
\newblock Intriguing properties of vision transformers.
\newblock \emph{NeurIPS}, 2021.

\bibitem[Niemeyer et~al.(2020)Niemeyer, Mescheder, Oechsle, and
  Geiger]{niemeyer2020differentiable}
Michael Niemeyer, Lars Mescheder, Michael Oechsle, and Andreas Geiger.
\newblock Differentiable volumetric rendering: Learning implicit 3d
  representations without 3d supervision.
\newblock In \emph{CVPR}, 2020.

\bibitem[Noroozi et~al.(2017)Noroozi, Pirsiavash, and
  Favaro]{noroozi2017representation}
Mehdi Noroozi, Hamed Pirsiavash, and Paolo Favaro.
\newblock Representation learning by learning to count.
\newblock In \emph{ICCV}, 2017.

\bibitem[Oord et~al.(2018)Oord, Li, and Vinyals]{oord2018representation}
Aaron van~den Oord, Yazhe Li, and Oriol Vinyals.
\newblock Representation learning with contrastive predictive coding.
\newblock \emph{arXiv:1807.03748}, 2018.

\bibitem[Oquab et~al.(2024)Oquab, Darcet, Moutakanni, Vo, Szafraniec, Khalidov,
  Fernandez, HAZIZA, Massa, El-Nouby, Assran, Ballas, Galuba, Howes, Huang, Li,
  Misra, Rabbat, Sharma, Synnaeve, Xu, Jegou, Mairal, Labatut, Joulin, and
  Bojanowski]{oquab2023dinov2}
Maxime Oquab, Timoth{\'e}e Darcet, Th{\'e}o Moutakanni, Huy~V. Vo, Marc
  Szafraniec, Vasil Khalidov, Pierre Fernandez, Daniel HAZIZA, Francisco Massa,
  Alaaeldin El-Nouby, Mido Assran, Nicolas Ballas, Wojciech Galuba, Russell
  Howes, Po-Yao Huang, Shang-Wen Li, Ishan Misra, Michael Rabbat, Vasu Sharma,
  Gabriel Synnaeve, Hu Xu, Herve Jegou, Julien Mairal, Patrick Labatut, Armand
  Joulin, and Piotr Bojanowski.
\newblock {DINO}v2: Learning robust visual features without supervision.
\newblock \emph{TMLR}, 2024.

\bibitem[Pan et~al.(2021)Pan, Dai, Liu, Loy, and Luo]{pan20212d}
Xingang Pan, Bo Dai, Ziwei Liu, Chen~Change Loy, and Ping Luo.
\newblock Do 2d gans know 3d shape? unsupervised 3d shape reconstruction from
  2d image gans.
\newblock 2021.

\bibitem[Pathak et~al.(2016)Pathak, Krahenbuhl, Donahue, Darrell, and
  Efros]{pathak2016context}
Deepak Pathak, Philipp Krahenbuhl, Jeff Donahue, Trevor Darrell, and Alexei~A
  Efros.
\newblock Context encoders: Feature learning by inpainting.
\newblock In \emph{CVPR}, 2016.

\bibitem[Pepik et~al.(2012)Pepik, Gehler, Stark, and Schiele]{pepik20123d}
Bojan Pepik, Peter Gehler, Michael Stark, and Bernt Schiele.
\newblock {3D2PM -- 3D Deformable Part Models}.
\newblock In \emph{ECCV}, 2012.

\bibitem[Rajasegaran et~al.(2022)Rajasegaran, Pavlakos, Kanazawa, and
  Malik]{rajasegaran2022tracking}
Jathushan Rajasegaran, Georgios Pavlakos, Angjoo Kanazawa, and Jitendra Malik.
\newblock Tracking people by predicting 3d appearance, location and pose.
\newblock In \emph{CVPR}, 2022.

\bibitem[Rajasegaran et~al.(2023)Rajasegaran, Pavlakos, Kanazawa,
  Feichtenhofer, and Malik]{rajasegaran2023benefits}
Jathushan Rajasegaran, Georgios Pavlakos, Angjoo Kanazawa, Christoph
  Feichtenhofer, and Jitendra Malik.
\newblock On the benefits of 3d pose and tracking for human action recognition.
\newblock In \emph{CVPR}, 2023.

\bibitem[Ranftl et~al.(2022)Ranftl, Lasinger, Hafner, Schindler, and
  Koltun]{Ranftl2022}
Ren\'{e} Ranftl, Katrin Lasinger, David Hafner, Konrad Schindler, and Vladlen
  Koltun.
\newblock Towards robust monocular depth estimation: Mixing datasets for
  zero-shot cross-dataset transfer.
\newblock \emph{PAMI}, 2022.

\bibitem[Russakovsky et~al.(2015)Russakovsky, Deng, Su, Krause, Satheesh, Ma,
  Huang, Karpathy, Khosla, Bernstein, et~al.]{russakovsky2015imagenet}
Olga Russakovsky, Jia Deng, Hao Su, Jonathan Krause, Sanjeev Satheesh, Sean Ma,
  Zhiheng Huang, Andrej Karpathy, Aditya Khosla, Michael Bernstein, et~al.
\newblock Imagenet large scale visual recognition challenge.
\newblock \emph{IJCV}, 2015.

\bibitem[Shen et~al.(2023)Shen, Yan, Qi, Najibi, Deng, Guibas, Zhou, and
  Anguelov]{shen2023gina}
Bokui Shen, Xinchen Yan, Charles~R Qi, Mahyar Najibi, Boyang Deng, Leonidas
  Guibas, Yin Zhou, and Dragomir Anguelov.
\newblock Gina-3d: Learning to generate implicit neural assets in the wild.
\newblock In \emph{CVPR}, 2023.

\bibitem[Shin et~al.(2018)Shin, Fowlkes, and Hoiem]{shin2018pixels}
Daeyun Shin, Charless~C Fowlkes, and Derek Hoiem.
\newblock Pixels, voxels, and views: A study of shape representations for
  single view 3d object shape prediction.
\newblock In \emph{CVPR}, 2018.

\bibitem[Silberman et~al.(2012)Silberman, Hoiem, Kohli, and
  Fergus]{silberman2012indoor}
Nathan Silberman, Derek Hoiem, Pushmeet Kohli, and Rob Fergus.
\newblock Indoor segmentation and support inference from rgbd images.
\newblock In \emph{ECCV}, 2012.

\bibitem[Skorokhodov et~al.(2023)Skorokhodov, Siarohin, Xu, Ren, Lee, Wonka,
  and Tulyakov]{3dgp}
Ivan Skorokhodov, Aliaksandr Siarohin, Yinghao Xu, Jian Ren, Hsin-Ying Lee,
  Peter Wonka, and Sergey Tulyakov.
\newblock 3d generation on imagenet.
\newblock In \emph{ICLR}, 2023.

\bibitem[Smith and Gasser(2005)]{smith2005development}
Linda Smith and Michael Gasser.
\newblock The development of embodied cognition: Six lessons from babies.
\newblock \emph{Artificial life}, 2005.

\bibitem[Spelke(1990)]{spelke1990principles}
Elizabeth~S Spelke.
\newblock Principles of object perception.
\newblock \emph{Cognitive science}, 1990.

\bibitem[Spelke and Kinzler(2007)]{spelke2007core}
Elizabeth~S Spelke and Katherine~D Kinzler.
\newblock Core knowledge.
\newblock \emph{Developmental science}, 2007.

\bibitem[Standley et~al.(2020)Standley, Zamir, Chen, Guibas, Malik, and
  Savarese]{standley2020tasks}
Trevor Standley, Amir Zamir, Dawn Chen, Leonidas Guibas, Jitendra Malik, and
  Silvio Savarese.
\newblock Which tasks should be learned together in multi-task learning?
\newblock In \emph{ICML}, 2020.

\bibitem[Tatarchenko et~al.(2019)Tatarchenko, Richter, Ranftl, Li, Koltun, and
  Brox]{tatarchenko2019single}
Maxim Tatarchenko, Stephan~R Richter, Ren{\'e} Ranftl, Zhuwen Li, Vladlen
  Koltun, and Thomas Brox.
\newblock What do single-view 3d reconstruction networks learn?
\newblock In \emph{CVPR}, 2019.

\bibitem[Van~Horn et~al.(2021)Van~Horn, Cole, Beery, Wilber, Belongie, and
  Mac~Aodha]{van2021benchmarking}
Grant Van~Horn, Elijah Cole, Sara Beery, Kimberly Wilber, Serge Belongie, and
  Oisin Mac~Aodha.
\newblock Benchmarking representation learning for natural world image
  collections.
\newblock In \emph{CVPR}, 2021.

\bibitem[Wang et~al.(2019)Wang, Ge, Lipton, and Xing]{wang2019learning}
Haohan Wang, Songwei Ge, Zachary Lipton, and Eric~P Xing.
\newblock Learning robust global representations by penalizing local predictive
  power.
\newblock In \emph{NeurIPS}, 2019.

\bibitem[Wu et~al.(2023)Wu, Li, Jakab, Rupprecht, and Vedaldi]{wu2023magicpony}
Shangzhe Wu, Ruining Li, Tomas Jakab, Christian Rupprecht, and Andrea Vedaldi.
\newblock Magicpony: Learning articulated 3d animals in the wild.
\newblock In \emph{CVPR}, 2023.

\bibitem[Wu et~al.(2018)Wu, Xiong, Yu, and Lin]{wu2018unsupervised}
Zhirong Wu, Yuanjun Xiong, Stella~X Yu, and Dahua Lin.
\newblock Unsupervised feature learning via non-parametric instance
  discrimination.
\newblock In \emph{CVPR}, 2018.

\bibitem[Yu et~al.(2021)Yu, Ye, Tancik, and Kanazawa]{yu2021pixelnerf}
Alex Yu, Vickie Ye, Matthew Tancik, and Angjoo Kanazawa.
\newblock pixelnerf: Neural radiance fields from one or few images.
\newblock In \emph{CVPR}, 2021.

\bibitem[Yu et~al.(2023)Yu, Xu, Zhang, Liu, Ye, Wu, Yan, Zhu, Xiong, Liang,
  et~al.]{yu2023mvimgnet}
Xianggang Yu, Mutian Xu, Yidan Zhang, Haolin Liu, Chongjie Ye, Yushuang Wu,
  Zizheng Yan, Chenming Zhu, Zhangyang Xiong, Tianyou Liang, et~al.
\newblock Mvimgnet: A large-scale dataset of multi-view images.
\newblock In \emph{CVPR}, 2023.

\bibitem[Zamir et~al.(2018)Zamir, Sax, Shen, Guibas, Malik, and
  Savarese]{zamir2018taskonomy}
Amir~R Zamir, Alexander Sax, William Shen, Leonidas~J Guibas, Jitendra Malik,
  and Silvio Savarese.
\newblock Taskonomy: Disentangling task transfer learning.
\newblock In \emph{CVPR}, 2018.

\bibitem[Zbontar et~al.(2021)Zbontar, Jing, Misra, LeCun, and
  Deny]{zbontar2021barlow}
Jure Zbontar, Li Jing, Ishan Misra, Yann LeCun, and St{\'e}phane Deny.
\newblock Barlow twins: Self-supervised learning via redundancy reduction.
\newblock In \emph{ICML}, 2021.

\bibitem[Zhang et~al.(2016)Zhang, Isola, and Efros]{zhang2016colorful}
Richard Zhang, Phillip Isola, and Alexei~A Efros.
\newblock Colorful image colorization.
\newblock In \emph{ECCV}, 2016.

\bibitem[Zhou et~al.(2022)Zhou, Wei, Wang, Shen, Xie, Yuille, and
  Kong]{zhou2021ibot}
Jinghao Zhou, Chen Wei, Huiyu Wang, Wei Shen, Cihang Xie, Alan Yuille, and Tao
  Kong.
\newblock ibot: Image bert pre-training with online tokenizer.
\newblock In \emph{ICLR}, 2022.

\bibitem[Zhou et~al.(2017)Zhou, Brown, Snavely, and Lowe]{zhou2017unsupervised}
Tinghui Zhou, Matthew Brown, Noah Snavely, and David~G Lowe.
\newblock Unsupervised learning of depth and ego-motion from video.
\newblock In \emph{CVPR}, 2017.

\end{thebibliography}
}

\end{document}